\documentclass[11pt]{article}
\usepackage{amssymb}
\usepackage{amsthm}
\usepackage{amsmath}
\usepackage{mathtools}
\usepackage{comment}
\usepackage{pdfpages}
\usepackage{float,booktabs}
\usepackage{threeparttable}
\usepackage{float} 
\usepackage{booktabs,tabularx,threeparttable}
\usepackage[table]{xcolor}
\usepackage{booktabs}
\usepackage{threeparttable}
\usepackage{array}
\usepackage{rotating}
\newcolumntype{L}[1]{>{\raggedright\arraybackslash}p{#1}}
\usepackage{graphicx}
\usepackage{hyperref}
\usepackage{natbib}
\usepackage{url}
\usepackage{xcolor}
\usepackage{dsfont}
\usepackage{soul,color,colortbl}
\usepackage{array,multirow}
\usepackage{makecell} 
\usepackage{cancel}
 
\usepackage[top=1in, bottom=1.3in, left=0.875in, right=0.875in]{geometry}
\usepackage[ruled,vlined]{algorithm2e}
\usepackage[labelformat=simple]{subcaption}
\usepackage{bm}
\usepackage{caption}

\def\spacingset#1{\renewcommand{\baselinestretch}
{#1}\small\normalsize} \spacingset{1}

\newcolumntype{P}[1]{>{\centering\arraybackslash}p{#1}}
\newcommand*{\myfont}{\fontfamily{lmss}\selectfont}
\DeclareTextFontCommand{\textpython}{\myfont}

\DeclareCaptionLabelFormat{boldnumber}{\bothIfFirst{#1}{~}\textbf{#2}}
\title{Joint Distributional Copula Regression for Insurance Claim Frequency and Severity via Boosting}

\author{Banghee So \thanks{Corresponding author: Department of Mathematics, Towson University, 7800 York Rd RM335, Towson, 21204, MD, USA. Email: \texttt{bso@towson.edu}.} 
\and Na Zhang \thanks{Department of Mathematics, Towson University, 7800 York Rd RM371, Towson, 21204, MD, USA. Email: \texttt{nzhang@towson.edu}.}}

\begin{document}

\maketitle

\begin{abstract}
In this paper, we introduce a model-based boosting approach for fitting structured additive distributional copula regression models for bivariate responses, with particular emphasis on jointly modeling insurance claim frequency and severity. The approach uses regression-type base learners, such as linear terms and splines with low degrees of freedom. It builds on generalized additive models for location, scale, and shape (GAMLSS), which extend generalized additive models (GAMs) by allowing multiple distributional parameters, rather than only the conditional mean, to depend on covariates. The structured additive predictors make the model inherently interpretable, offering greater transparency and auditability than black-box approaches. These features are particularly relevant in regulated actuarial settings and applications requiring explainable artificial intelligence. By estimating the entire conditional joint distribution, the framework also enables insurance practitioners to examine conditional quantiles, tail behavior, and prediction intervals of risks. The copula component captures dependence between claim frequency and severity and allows practitioners to investigate how this dependence varies with policyholder characteristics, potentially improving predictive performance. Moreover, by incorporating modern machine-learning techniques, specifically boosting, the proposed model improves prediction accuracy while retaining interpretability. We assess the proposed framework using both simulated and real data. The results demonstrate its ability to recover the true covariate effects on the marginal distributional parameters and the dependence structure, select informative variables, and achieve predictive performance competitive with standard benchmark models. We also illustrate how to interpret the fitted model.
\end{abstract}

\newpage
\section{Introduction} \label{sec:intro}
While modeling the conditional mean of a response is sufficient for some applications, many require a deeper understanding of the entire conditional distribution, including its quantiles, tail behavior, and prediction intervals. Consequently, distributional regression models, which allow covariates to affect the entire response distribution rather than only its mean, have received increasing attention in statistical research over the past decade \citep{klein2024distributional}. This broader perspective is particularly important in actuarial applications, where effective insurance pricing and risk management demand an understanding of how policyholder characteristics drive not only expected claims but also claim variability and tail risk. Although actuarial research has increasingly considered the full conditional distribution, the literature remains relatively limited \citep{avanzi2024ensemble, pala2026agricultural}.

A second modeling challenge concerns the dependence between insurance outcomes. Claim frequency and claim severity, for example, may be driven by common risk factors and therefore be dependent. Nevertheless, traditional approaches often model them separately, potentially overlooking this dependence. Copula models offer a flexible solution by linking suitable marginal models through a dependence structure. Copula-based regression has consequently been used in several actuarial studies to jointly model insurance outcomes \citep{shi2020regression, kularatne2021use, power2024flexible}.

At the same time, advances in machine learning (ML) and deep learning (DL) have intensified the trade-off between predictive accuracy and model interpretability. Since their introduction by \citep{nelder1972generalized}, generalized linear models (GLMs) have played a central role in non-life insurance pricing because of their transparent parameterization and consistency with actuarial practice and regulatory expectations for interpretability. However, standard GLMs rely on prespecified predictor structures and distributional assumptions that may fail to capture complex nonlinear relationships, thereby limiting their predictive performance. Recent studies have shown that ML and DL methods can improve predictive accuracy by capturing such complex effects in high-dimensional data \citep{wuthrich2020bias, agarwal2021neural, arik2021tabnet, mcdonnell2023deep}. In particular, boosting methods such as Extreme Gradient Boosting, LightGBM, and CatBoost have produced promising results in automobile insurance pricing, including applications involving high-dimensional telematics data \citep{so2021cost,henckaerts2021boosting, so2024enhanced, power2024flexible, so2025zero}.
These improvements in predictive performance often come at the cost of reduced transparency. Many modern ML and DL methods operate as black-box models whose predictions are difficult to audit or justify, creating practical challenges in regulated insurance settings \citep{mamiko2020impact}. As in other highly regulated industries, insurance models are increasingly expected to satisfy standards of explainability, fairness, and transparency \citep{henckaerts2022stakes}. Post-hoc explanation approaches, including LIME \citep{ribeiro2016should} and SHAP \citep{lundberg2017unified}, can yield informative insights into individual predictions and the contributions of individual variables. Nevertheless, these techniques do not render the underlying model inherently interpretable, and the resulting explanations can be sensitive to the chosen approximation procedure or to minor variations in the fitted model \citep{alvarez2018robustness}. These limitations have motivated the development of intrinsically interpretable, or glass-box, models within the broader field of explainable artificial intelligence \citep{ali2023explainable}. Generalized additive models (GAMs; \citep{hastie1986generalized}) allow nonlinear feature effects while preserving a transparent additive structure. More recently, \citet{krupova2026explainable} studied Explainable Boosting Machines, which combine the additive structure of GAMs with flexible boosting, while \citet{richman2023localglmnet} introduced LocalGLMnet, which retains a GLM-like representation while allowing regression coefficients to be learned through a neural network. 

Motivated by these three imperatives--distributional flexibility, joint outcome dependence, and intrinsic interpretability--we propose a model-based boosting approach for structured additive distributional copula regression. Our primary focus is the joint modeling of claim frequency and severity, representing a mixed-response scenario where frequency is a discrete count and severity is a continuous variable. Building on the generalized additive models for location, scale, and shape (GAMLSS; \citealp{rigby2005generalized}) framework, our approach links each parameter of the two marginal distributions, as well as the copula dependence parameter, to a structured additive predictor. Consequently, policyholder characteristics can simultaneously influence both the marginal distributions and their dependence structure. Estimation is performed using component-wise model-based boosting. Model-based boosting combines functional gradient descent \citep{friedman2001greedy} with the component-wise fitting of predefined, regression-like base learners—such as simple linear models or splines with low degrees of freedom \citep{buhlmann2003boosting}. \citet{mayr2012generalized} subsequently extended this approach to GAMLSS. At each boosting iteration, the algorithm updates the base learner that best fits the current negative gradient, thereby providing an intrinsic mechanism for variable selection. This procedure can accommodate high-dimensional settings, including cases in which the number of covariates exceeds the number of observations. Moreover, small step lengths and early stopping regularize the fitted effects by shrinking them toward zero, which can reduce estimation variability and improve out-of-sample predictive performance \citep{mayr2012generalized}. Model-based boosting, therefore, provides a useful alternative to conventional unpenalized maximum likelihood estimation and complements existing Bayesian and likelihood approaches to copula regression.

The proposed framework addresses multiple actuarial modeling issues at once. First, it estimates the full conditional joint distribution of the outcomes instead of restricting attention to conditional means, enabling a richer understanding of the distribution as a whole, including quantiles, tail properties, and prediction intervals. Second, the model accounts for dependence between claim frequency and severity, allowing practitioners to study how this dependence changes with policyholder characteristics and potentially enhancing predictive accuracy. It also performs regularized variable selection in potentially high-dimensional settings, so when the data contain many irrelevant covariates, the model automatically identifies and retains the informative ones. Moreover, the use of structured additive predictors makes it possible to assess how each covariate influences both the marginal parameters and the dependence parameters. Consequently, the approach remains interpretable via its additive components, providing greater transparency and auditability than black-box methods—an important advantage in regulated actuarial environments and other applications requiring explainable AI. Finally, by incorporating modern machine learning techniques, in particular boosting, the framework achieves improved predictive performance relative to traditional models while retaining interpretability. We assess the framework using both simulated and real-world data. In simulation studies, we evaluate its ability to recover the true covariate effects and dependence structure, to select relevant variables, and to deliver strong predictive performance compared with benchmark approaches. In an empirical application based on telematics data, we illustrate how the framework models the joint distribution of claim frequency and severity and compare its predictive accuracy with that of separate modeling strategies. We also examine the estimated covariate effects to show how the fitted model can be interpreted.

The remainder of the paper is organized as follows. Section~\ref{sec:sc2} reviews the framework for bivariate distributional copula regression with mixed discrete and continuous responses. Section~\ref{sec:sc3} details the model-based boosting approach. Section~\ref{sec:sc4} presents the simulation studies and summarizes their results. Section~\ref{sec:sc5} illustrates the practical utility of our framework through an empirical analysis using real-world insurance telematics data. Finally, Section~\ref{sec:conclude} concludes the paper.

\section{Joint distributional copula regression} \label{sec:sc2}
We provide an overview of structured additive distributional copula regression, with an emphasis on the modeling of bivariate responses in both discrete and continuous settings.

\subsection{Model structure} \label{subsec:model-structure}
In the bivariate response setting, let \((Y_1,Y_2)\) denote a bivariate response associated with a covariate vector \(\bm{x}\). We assume that the conditional marginal distributions of \(Y_1\) and \(Y_2\) belong to parametric families with cumulative distribution functions (CDFs)
\[F_1\left(y_1\mid\bm{\theta}^{(1)}\right)
\quad\text{and}\quad
F_2\left(y_2\mid\bm{\theta}^{(2)}\right),\]
respectively, where
\[\bm{\theta}^{(1)}=
\left(\theta_1^{(1)},\ldots,\theta_{K_1}^{(1)}\right)^\top\]
and
\[\bm{\theta}^{(2)}=
\left(\theta_1^{(2)},\ldots,\theta_{K_2}^{(2)}\right)^\top\]
are the corresponding marginal parameter vectors. Here, \(K_1\) and \(K_2\) denote the numbers of parameters in the first and second marginal distributions, respectively. 

Following Sklar's theorem \citep{sklar1959fonctions}, the joint conditional CDF can be expressed as
\begin{equation}
\label{eq:copula-rep}
F\left(y_1,y_2\mid\bm{\theta}\right)
=C\left(
F_1\left(y_1\mid\bm{\theta}^{(1)}\right),
F_2\left(y_2\mid\bm{\theta}^{(2)}\right)\mid
\theta^{(c)}
\right),
\end{equation}
where \(C:[0,1]^2\rightarrow[0,1]\) is a bivariate copula and \(\theta^{(c)}\) is its scalar dependence parameter. The copula separates the marginal distributions from their dependence structure, allowing them to be specified independently. When both margins are continuous, the copula in Equation~\eqref{eq:copula-rep} is unique. If at least one margin is discrete, it is unique only over the ranges of the marginal CDFs. Nevertheless, the copula representation still defines a valid joint distribution. For a discrete--continuous response pair, the likelihood contribution is obtained by taking a finite difference with respect to the discrete margin and differentiating with respect to the continuous margin, as described in the following subsection. Let $K=K_1+K_2+1$ denote the total number of parameters, including the copula parameter. The full parameter vector is
$\bm{\theta}=\left(\left(\bm{\theta}^{(1)}\right)^\top,
\left(\bm{\theta}^{(2)}\right)^\top,\theta^{(c)}\right)^\top.$

In the joint distributional copula regression framework, each component of \(\bm{\theta}\) is related to the covariates through a parameter-specific structured additive predictor. For notational convenience, let \(K_c=1\) and \(\theta_1^{(c)}=\theta^{(c)}\). For \(\nu\in\{1,2,c\}\) and \(k=1,\ldots,K_\nu\), the predictor is defined as
\begin{equation}
\label{eq:additive-predictor}
\eta_k^{(\nu)}(\bm{x})
=\beta_{0k}^{(\nu)}+\sum_{r=1}^{P_k^{(\nu)}}s_{rk}^{(\nu)}(x_r)
=g_k^{(\nu)}\left(\theta_k^{(\nu)}(\bm{x})\right).
\end{equation}
Here, \(\beta_{0k}^{(\nu)}\) is an intercept, \(g_k^{(\nu)}(\cdot)\) is a suitable link function, and \(s_{rk}^{(\nu)}(\cdot)\) represents the effect of the \(r\)th covariate. The quantity \(P_k^{(\nu)}\) denotes the number of covariate effects included in the predictor. It may differ across parameters, allowing each marginal and dependence parameter to be modeled using a different subset of covariates.

In Equation~\eqref{eq:additive-predictor}, a linear covariate effect is given by
\[s_{rk}^{(\nu)}(x_r)=\beta_{rk}^{(\nu)}x_r,\]
where \(\beta_{rk}^{(\nu)}\) is the corresponding regression coefficient. A nonlinear effect can be represented using B-spline basis functions:
\[s_{rk}^{(\nu)}(x_r)=\sum_{\ell=1}^{L_r}\beta_{rk\ell}^{(\nu)}
B_{r\ell}(x_r),\]
where \(B_{r\ell}(\cdot)\) is the \(\ell\)th B-spline basis function and \(\beta_{rk\ell}^{(\nu)}\) is its coefficient.

The corresponding marginal or copula parameter is obtained through the inverse link function:
\[
\theta_k^{(\nu)}(\bm{x})
=
\left(g_k^{(\nu)}\right)^{-1}
\left(
\eta_k^{(\nu)}(\bm{x})
\right).
\]
The link function is chosen to ensure that each parameter remains within its admissible range. 

Although the framework can accommodate a broad range of copula families, we consider the Gumbel and Frank Archimedean copulas, motivated by their usefulness in insurance dependence modeling \citep{kularatne2021use}, together with the widely used Gaussian copula. The Gumbel copula captures positive dependence with upper-tail dependence, whereas the Frank copula captures symmetric positive or negative dependence without asymptotic tail dependence. The Gaussian copula provides a symmetric, tail-independent benchmark. These families therefore allow us to examine the proposed framework under distinct dependence structures.

\subsection{Application to discrete--continuous responses}
\label{subsec:log-likelihood}
We now specialize the general framework in Section~\ref{subsec:model-structure} to a mixed discrete--continuous response. We assume that \(Y_1\) is discrete and \(Y_2\) is continuous. For notational simplicity, define
\[F_1=F_1\left(y_1\mid\bm{\theta}^{(1)}\right),\qquad
F_2=F_2\left(y_2\mid\bm{\theta}^{(2)}\right),\]
and let
\[f_1=f_1\left(y_1\mid\bm{\theta}^{(1)}\right)
:=\mathbb{P}\left(Y_1=y_1\mid\bm{\theta}^{(1)}\right),\qquad
f_2=f_2\left(y_2\mid\bm{\theta}^{(2)}\right),\]
denote the marginal probability mass and density evaluated at the observed responses, respectively. Since \(Y_1\) is discrete, we further define
\[F_1^{-}:=F_1-f_1=\mathbb{P}\left(Y_1<y_1\mid\bm{\theta}^{(1)}
\right),\]
which is the value of the marginal CDF immediately before its jump at \(y_1\).
By the copula representation in Equation~\eqref{eq:copula-rep},
\[\mathbb{P}\left(Y_1\leq y_1,\,Y_2\leq y_2\mid\bm{\theta}\right)
=C\left(F_1,F_2 \mid \theta^{(c)}\right).\]
Because \(Y_1\) is discrete, the probability associated with the event \(Y_1=y_1\) is obtained by taking a finite difference with respect to the first margin:
\begin{align*}
\mathbb{P}\left(
Y_1=y_1,\,
Y_2\leq y_2
\mid\bm{\theta}
\right)
&=
\mathbb{P}\left(
Y_1\leq y_1,\,
Y_2\leq y_2
\mid\bm{\theta}
\right)
-
\mathbb{P}\left(
Y_1<y_1,\,
Y_2\leq y_2
\mid\bm{\theta}
\right)
\\
&=
C\left(
F_1,F_2\mid\theta^{(c)}
\right)
-
C\left(
F_1^{-},F_2\mid\theta^{(c)}
\right).
\end{align*}
Since \(Y_2\) is continuous, the mixed probability-mass/density contribution is obtained by differentiating this expression with respect to \(y_2\). Specifically, define \(p(y_1,y_2\mid\bm{\theta})\) such that
\[\mathbb{P}\left(Y_1=y_1,\,Y_2\in dy_2\mid\bm{\theta}\right)
=p\left(y_1,y_2\mid\bm{\theta}\right)\,dy_2.\]
For notational convenience, let
\begin{equation}
\label{eq:C2}
C_2(u,v\mid\theta)=\frac{\partial C(u,v\mid\theta)}{\partial v}
\end{equation}
denote the partial derivative of the copula with respect to its second argument. Applying the chain rule gives
\[p\left(y_1,y_2\mid\bm{\theta}\right)=
\left[C_2\left(F_1,F_2\mid\theta^{(c)}\right)-
C_2\left(F_1^{-},F_2\mid\theta^{(c)}\right)\right]f_2.\]
Therefore, the contribution of a generic observation to the log-likelihood is
\begin{equation}
\label{eq:log-likelihood}
\ell\left(y_1,y_2\mid\bm{\theta}\right)=
\log\left[C_2\left(F_1,F_2\mid\theta^{(c)}\right)-
C_2\left(F_1^{-},F_2\mid\theta^{(c)}\right)\right]+\log f_2.
\end{equation}
For \(n\) independent observations indexed by \(i\), we obtain the joint log-likelihood
\begin{equation}
\label{eq:joint-log-likelihood}
\begin{aligned}
\ell\left(\bm{\theta}_{1:n}\right)
&=
\sum_{i=1}^{n}
\ell\left(y_{1i},y_{2i}\mid \bm{\theta}_i\right)
\\
&=
\sum_{i=1}^{n}
\left\{
\log\left[
C_2\left(
F_{1i},F_{2i}\mid\theta_i^{(c)}
\right)
-
C_2\left(
F_{1i}^{-},F_{2i}\mid\theta_i^{(c)}
\right)
\right]
+
\log f_{2i}
\right\},
\\[-0.2em]
\bm{\theta}_{1:n}
&=
\left(
\bm{\theta}_1^\top,\ldots,\bm{\theta}_n^\top
\right)^\top.
\end{aligned}
\end{equation}

This expression applies to any copula for which the partial derivative with respect to its second argument exists. Appendix~A provides the corresponding log-likelihood for Poisson--Gamma margins coupled through a Gaussian copula.

\section{Estimation via model-based boosting} \label{sec:sc3}
Our approach proposes estimating distributional copula regression models using model-based boosting. In this section, we give an overview of the algorithm.

\subsection{Boosting algorithm for joint distributional copula regression} \label{subsec:boosting}
The parameter estimation for the distributional copula regression model with a discrete–continuous response \(\bm Y=(Y_{1},Y_{2})^\top\) can be formulated as the following optimization problem.
\[\widehat{\bm{\eta}}=
\operatorname*{arg\,min}_{\bm{\eta}}
\left[\mathbb{E}_{\bm{Y}}
\left\{\omega\left(
\bm{Y}\mid \bm{\eta} \right)\right\}\right],\]
where
\[\bm{\eta}=\left(\eta_1^{(1)},\ldots,\eta_{K_1}^{(1)},
\eta_1^{(2)},\ldots,\eta_{K_2}^{(2)},\eta^{(c)}\right)^\top
\in\mathbb{R}^K\]
is the vector of additive predictors associated with the model parameters, with each component specified in Equation~\eqref{eq:additive-predictor}. The corresponding estimates of these additive predictor functions are given by
\[\widehat{\bm{\eta}}=\left(
\widehat{\eta}_1^{(1)},\ldots,\widehat{\eta}_{K_1}^{(1)},
\widehat{\eta}_1^{(2)},\ldots,\widehat{\eta}_{K_2}^{(2)},
\widehat{\eta}^{(c)}\right)^\top\in\mathbb{R}^K,\]
$\omega(\cdot)$ represents the loss function. For observations $i=1,\ldots,n$, we minimize the empirical risk
\[\frac{1}{n}\sum_{i=1}^{n}
\omega\left(\bm{y}_i \mid \bm{\eta}_i\right)\] instead. A common choice for $\omega$ is the negative log-likelihood; accordingly, in this paper we adopt the negative log-likelihood of the bivariate response distribution in Equation~\eqref{eq:joint-log-likelihood} as the loss function.

\begin{figure}[htbp]
    \centering
    \includegraphics[width=1\textwidth]{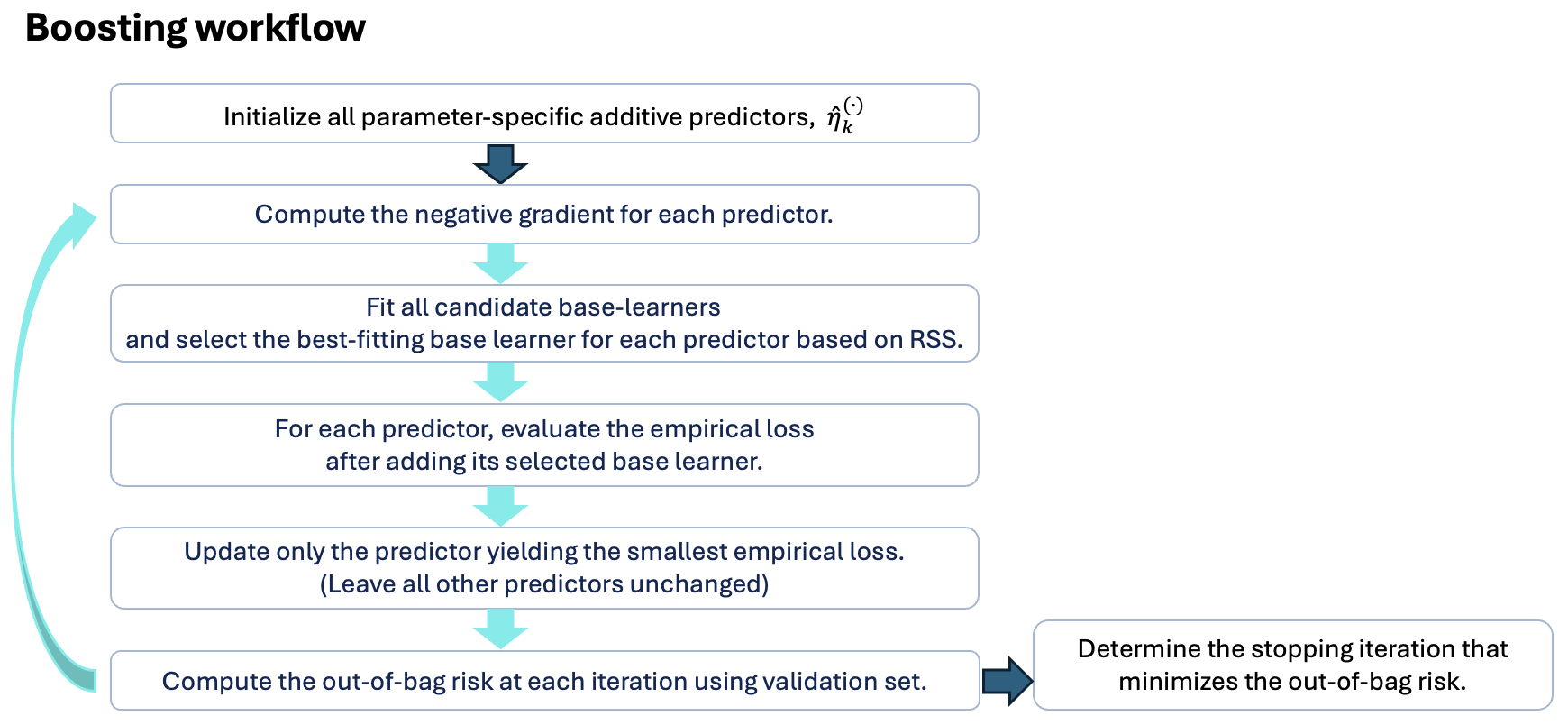}
    \caption{Noncyclic model-based boosting algorithm}
    \label{fig:boosting algorithm}
\end{figure}

The fundamental idea behind model-based gradient boosting is to iteratively minimize the empirical risk through a stagewise descent that uses the gradient of the loss function. Similar to the classical boosting algorithm, it is a sequential ensemble approach in which the ensemble is composed of regression-function base learners. More specifically, the algorithm starts with initial values for the additive predictors and with a prespecified set of regression-type base learners for each parameter. Common options include simple linear models and regression splines with only a few degrees of freedom. The chosen base-learner type determines the kind of contribution a covariate makes to the associated additive predictor, for instance a linear versus a nonlinear effect. At iteration $m$, parameter-specific base learners are fitted to the corresponding negative partial gradients computed at the estimates from iteration $m-1$. The next subsection presents these negative gradients for each parameter for a discrete--continuous response. These gradient vectors may be viewed as pseudo-observations; for squared-error loss, they are identical to the residuals from the previous boosting step. Here, they are obtained from the first derivatives of the negative log-likelihood with respect to the associated parameter predictors $\eta_k^{(\bullet)}$. Next, for each parameter-specific gradient, the base learner providing the best fit according to a residual sum-of-squares criterion is selected. Each selected base learner is then temporarily added to its corresponding predictor, and the updated empirical loss is computed. From these candidate updates, only the predictor that achieves the lowest empirical loss is actually updated in a given iteration; this is known as noncyclic boosting. Since updates are applied only to the selected base learners, the method also performs inherent variable selection.

Two algorithmic approaches can be used to boost distributional regression models: the cyclic algorithm \citep{mayr2012generalized} and the noncyclic algorithm \citep{thomas2017probing}. In the cyclic approach, the additive predictors are updated sequentially by fitting the best-performing base learner for each predictor in turn. The noncyclic approach adds an extra selection step across the different model parameters. Concretely, at each iteration, only the parameter-specific predictor that yields the largest overall reduction in loss is updated, using its best-fitting base learner. Due to its advantages in variable selection, we focus on the noncyclic algorithm in what follows. The algorithm is given in Appendix B and is briefly outlined in Figure~\ref{fig:boosting algorithm}. In model-based boosting, the optimal number of iterations can be determined by minimizing the out-of-bag risk (\citet{buhlmann2007boosting, mayr2012generalized}). To this end, the complete dataset is split into a training set and a validation set. The boosting algorithm is fitted on the training data, and its predictive performance is assessed on the validation data using the risk. The stopping iteration is then selected as the iteration at which the out-of-bag risk reaches its minimum.

\subsection{Negative gradients of the implemented loss function}\label{subsec:negative-gradients}
Recall that the loss is defined as the negative log-likelihood, i.e., $\omega=-\ell$. Hence, taking the negative derivative of the loss with respect to an additive predictor yields the associated partial derivative of the log-likelihood:
\[-\frac{\partial\omega}{\partial\eta_k^{(\nu)}} =
\frac{\partial\ell}{\partial\eta_k^{(\nu)}},\qquad
\nu\in\{1,2\},\quad k=1,\ldots,K_\nu.\]
For the copula parameter, the relevant additive predictor is written as $\eta^{(c)}$.

To simplify notation, define
\begin{equation}
\label{eq:A}
A=C_2\left(F_1,F_2\mid\theta^{(c)}\right)-
C_2\left(F_1^{-},F_2\mid\theta^{(c)}\right).
\end{equation}
The log-likelihood in Equation~\eqref{eq:log-likelihood} can then be written as
\[\ell=\log A+\log f_2.\]
We further define the following partial derivatives of $C_2$:
\begin{equation}
\label{eq:C21-C22}
\begin{aligned}
C_{21}(u,v\mid\theta)
&=\frac{\partial C_2(u,v\mid\theta)}{\partial u},\\
C_{22}(u,v\mid\theta)
&=\frac{\partial C_2(u,v\mid\theta)}{\partial v},\\
C_{2\theta}(u,v\mid\theta)
&=\frac{\partial C_2(u,v\mid\theta)}{\partial\theta}.
\end{aligned}
\end{equation}
For a parameter of the first marginal distribution, the negative
gradient is
\begin{equation}
\label{eq:negative-gradient-margin1}
\begin{aligned}
\frac{\partial\ell}{\partial\eta_k^{(1)}}=\frac{1}{A}\Bigg[
&C_{21}\left(F_1,F_2\mid\theta^{(c)}\right)
\frac{\partial F_1}{\partial\eta_k^{(1)}}\\
&-C_{21}\left(F_1^{-},F_2\mid\theta^{(c)}\right)
\left(\frac{\partial F_1}{\partial\eta_k^{(1)}}-
\frac{\partial f_1}{\partial\eta_k^{(1)}}\right)\Bigg],\qquad k=1,\ldots,K_1.
\end{aligned}
\end{equation}
For a parameter of the second marginal distribution, the negative gradient is
\begin{equation}
\label{eq:negative-gradient-margin2}
\begin{aligned}
\frac{\partial\ell}{\partial\eta_k^{(2)}}={}&
\frac{1}{A}\left[C_{22}\left(F_1,F_2\mid\theta^{(c)}\right)
-C_{22}\left(F_1^{-},F_2\mid\theta^{(c)}\right)\right]
\frac{\partial F_2}{\partial\eta_k^{(2)}}\\&+
\frac{1}{f_2}\frac{\partial f_2}{\partial\eta_k^{(2)}},
\qquad k=1,\ldots,K_2.
\end{aligned}
\end{equation}
Finally, the negative gradient with respect to the copula predictor is
\begin{equation}
\label{eq:negative-gradient-copula}
\frac{\partial\ell}{\partial\eta^{(c)}}=
\frac{1}{A}\left[
C_{2\theta}\left(F_1,F_2\mid\theta^{(c)}\right)-
C_{2\theta}\left(F_1^{-},F_2\mid\theta^{(c)}\right)\right]
\frac{\partial\theta^{(c)}}{\partial\eta^{(c)}}.
\end{equation}
\section{Simulated experiment} \label{sec:sc4}
In this section, we evaluate the proposed distributional copula regression framework fitted using model-based boosting through a simulation study in which the model parameters and data-generating mechanism are fully specified. The main objectives are to assess (i) predictive performance relative to benchmark approaches that fit the two responses separately and therefore assume independence, (ii) the recovery of the underlying covariate effects and dependence structure, and (iii) variable-selection performance.

\subsection{Data generation} \label{subsec:synthetic data generation}
For the discrete--continuous response setting, we generate the discrete margin from a Poisson distribution and the continuous margin from a Gamma distribution. The two margins are linked using a Frank copula. Although the Frank copula can accommodate both positive and negative dependence, the copula predictor specified below takes only positive values and therefore generates positive dependence in this simulation. All covariates are generated independently from a uniform distribution on \([0,1]\). To assess the ability of the proposed method to recover different forms of covariate effects, the distributional predictors include both linear and nonlinear effects:
\begin{equation}
\label{eq:simulation-predictors}
\begin{aligned}
\ln(\lambda) &=\eta_1^{(1)}=\frac{1}{2}
\left(X_1^{3/2}-2\cos(3X_1)\right) +1.3X_2-X_3+2X_4 
&& \text{(Poisson mean)}, \\[1ex]
\ln(\mu) &=\eta_1^{(2)}= 1.5X_2+0.5X_3+5X_4^3-6X_4^2+2X_4
&& \text{(Gamma mean)}, \\[1ex]
\ln(\sigma) &=\eta_2^{(2)}=
-1.5\left(1.5\cos(2X_5)+3\tanh(X_5)\right)+2X_6, 
&&\text{(Gamma dispersion)}\\[1ex]
\theta^{(c)} &=\eta^{(c)}=2\sin(4X_7)+3 &&\text{(Frank copula parameter)}.
\end{aligned}
\end{equation}

To evaluate the variable-selection performance of the boosting algorithm, we consider three covariate-dimensional settings with \(p\in\{10,100,1000\}\). In each setting, the seven covariates \(X_1,\ldots,X_7\) are informative for at least one distributional or copula parameter, as specified in Equation~\eqref{eq:simulation-predictors}, while the remaining \(p-7\) covariates are noise variables. Thus, the three settings contain 3, 93, and 993 noise variables, respectively. For each simulated data set, we create 2,000 observations. The first 1,000 observations are used as the training set to fit the model. We then use 500 observations as a tuning set to choose the best stopping iteration, and the final 500 observations serve as the test set for evaluating out-of-sample performance. For each dimensional setting, we generate 10 independent data sets and report the average and standard deviation of each performance metric over these 10 runs.

\subsection{Estimation and evaluation setup} \label{subsec:synthetic setting}
Since the functional form of each covariate effect is generally unknown in practice, we specify a B-spline base learner with three effective degrees of freedom for every covariate in each distributional predictor. This flexible specification allows both linear and nonlinear effects to be recovered without prespecifying their functional forms. The step length is fixed at \(s_{\mathrm{step}}=0.1\) for all distributional parameters. The maximum number of boosting iterations is set to \(m_{\mathrm{stop}}=5{,}000\), and the optimal stopping iteration is selected by minimizing the out-of-bag risk evaluated on the tuning sample.

We compare the proposed model with two benchmark approaches. The first benchmark consists of separate univariate distributional models for the Poisson and Gamma margins, both fitted using model-based boosting. Because the two margins are modeled separately, this approach assumes independence between the responses. The second benchmark also fits separate Poisson and Gamma models under the independence assumption, but estimates them by maximum likelihood rather than boosting. 

We evaluate the predictive performance of the proposed copula model and the two benchmark approaches using two proper scoring rules: the negative log-likelihood and the energy score \citep{gneiting2008assessing}. For the independent benchmark models, the joint negative log-likelihood is calculated as the sum of the negative marginal log-likelihoods. Let \(\bm{y}=(y_1,y_2)^\top\) denote an observed bivariate outcome and let \(\widehat{F}_Y\) denote its predictive distribution. The energy score is defined as:
\begin{equation}
\label{eq:energy-score}
\widehat{\operatorname{ES}}
\left(\widehat{F}_Y,\bm{y}\right)
=
\frac{1}{M}
\sum_{m=1}^{M}
\left\|\bm{X}_m-\bm{y}\right\|
-
\frac{1}{2M^2}
\sum_{m=1}^{M}
\sum_{m'=1}^{M}
\left\|\bm{X}_m-\bm{X}_{m'}\right\|,
\end{equation}
where \(\|\cdot\|\) denotes the Euclidean norm on \(\mathbb{R}^2\), and \(M\) predictive samples \(\bm{X}_1,\ldots,\bm{X}_M\) are drawn independently from \(\widehat{F}_Y\)
with
\[
\bm{X}_m
=
(X_{m1},X_{m2})^\top
\in\mathbb{R}^2,
\qquad m=1,\ldots,M.
\]
The first term measures the distance between the predictive samples and the observed outcome, whereas the second term accounts for the dispersion of the predictive distribution. The energy score therefore evaluates the entire joint predictive distribution, including its marginal behavior and dependence structure. As with the negative log-likelihood, a lower energy score indicates better predictive performance.

\subsection{Results} \label{subsec:simulation result}
We evaluate the proposed method along three dimensions: predictive performance, recovery of the underlying covariate effects and dependence structure, and variable-selection performance.

\noindent\textbf{Predictive performance.} 
As expected, Table~\ref{tab:performance_summary} indicates that the Frank copula boosting model attains the smallest negative log-likelihood and energy score. The model is fitted to the simulated data using a Frank copula distributional regression approach estimated via noncyclic model-based boosting. Reported values are means across 10 independent replications, with standard deviations given in parentheses. To compute the energy score, 1{,}000 samples ($M=1{,}000$) were generated.

\begin{table}[htbp]
\centering
\caption{Predictive performance of the proposed model and two benchmark models across different dimensional settings}
\label{tab:performance_summary}
\small
\renewcommand{\arraystretch}{1.2}
\begin{tabular}{p{2.6cm} l c c c}
\toprule
\multirow{2}{*}{\textbf{Score}}
& \multirow{2}{*}{\textbf{Model}}
& \multicolumn{3}{c}{\textbf{Mean (SD)}} \\
\cmidrule(lr){3-5}
& & \(\bm{p=10}\) & \(\bm{p=100}\) & \(\bm{p=1000}\) \\
\midrule

\multirow{3}{2.6cm}{Negative log-likelihood}
& Frank copula boosting
    & \color{red}1,293 (24) & \color{red}1,344 (27) & \color{red}1,386 (31) \\
& Univariate boosting
    & 1,367 (29) & 1,389 (28) & 1,411 (31) \\
& Univariate MLE
    & 1,396 (29) & 25,626 (19,180) & -- \\
\midrule

\multirow{3}{2.6cm}{Energy score}
& Frank copula boosting
    & \color{red}1.40 (0.04) & \color{red}1.44 (0.04) &\color{red} 1.47 (0.03) \\
& Univariate boosting
    & 1.41 (0.05) & 1.45 (0.04) & 1.48 (0.03) \\
& Univariate MLE
    & 1.43 (0.06) & 2.05 (0.10) & -- \\
\bottomrule
\end{tabular}
\begin{minipage}{0.95\textwidth}
\footnotesize
\textit{Note:} Entries report the mean, with the standard deviation in
parentheses, over 10 data sets. The univariate MLE
model could not be estimated for \(p=1000\).
\end{minipage}
\end{table}

Across all three dimensional settings, the Frank copula boosting model achieves the lowest negative log-likelihood and energy score, followed by univariate boosting. This result indicates that accounting for the dependence between the two responses improves the estimation of their joint predictive distribution. The performance of both boosting approaches deteriorates moderately as the number of covariates increases from \(p=10\) to \(p=1{,}000\). In contrast, the univariate MLE model becomes unstable as the dimensionality increases. For \(p=100\), it produces substantially larger negative log-likelihood and energy score values than the boosting models, and it cannot be fitted successfully for \(p=1{,}000\). These findings demonstrate the greater stability of model-based boosting in the settings considered here. In particular, boosting remains feasible when the number of candidate covariates equals the training sample size, whereas conventional maximum likelihood estimation encounters severe numerical and estimation difficulties.

To assess the model's sensitivity to the choice of copula family, we additionally estimate the proposed framework using Gaussian and Gumbel copulas. Given that the synthetic data are generated from a Frank copula, these alternatives represent misspecified dependence structures. Table~\ref{tab:performance_summary_p10} compares their predictive performance against the correctly specified Frank copula model for \(p=10\).
\begin{table}[htbp]
\centering
\caption{Effect of copula misspecification on predictive performance for \(p=10\)}
\label{tab:performance_summary_p10}
\small
\renewcommand{\arraystretch}{1.2}
\begin{tabular}{lcc}
\toprule
\textbf{Model}
& \textbf{Negative log-likelihood}
& \textbf{Energy score} \\
\midrule
Frank copula boosting
    & \color{red}1,293 (24) & \color{red}1.40 (0.04) \\
Gaussian copula boosting
    & 1,314 (26) & 1.41 (0.04) \\
Gumbel copula boosting
    & 1,323 (32) & 1.41 (0.05) \\
\bottomrule
\end{tabular}
\end{table}
The correctly specified Frank copula model achieves the best predictive accuracy, yielding the lowest negative log-likelihood and energy score. Notably, despite their misspecified dependence structures, both the Gaussian and Gumbel copula models outperform the independence-based benchmarks (shown in Table~\ref{tab:performance_summary}) in terms of negative log-likelihood. Furthermore, their energy scores remain competitive with the univariate boosting and MLE model. These findings demonstrate that explicitly modeling dependence can improve joint probabilistic prediction even when the copula family is not correctly specified, although selecting an appropriate copula family provides an additional improvement.

\begin{figure}[htbp]
    \centering
    \includegraphics[width=\textwidth]{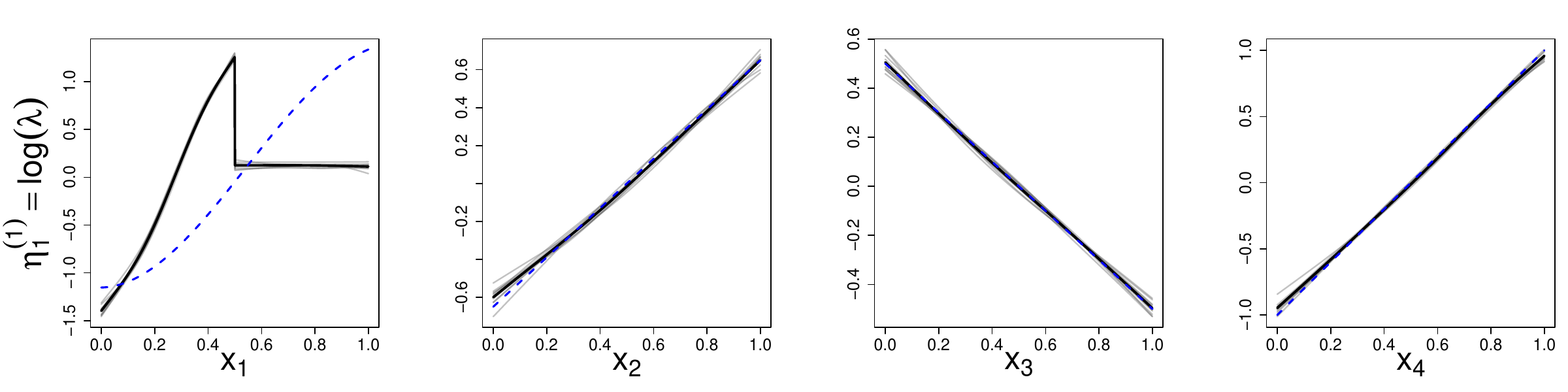}
    \par\medskip
    \includegraphics[width=0.75\textwidth]{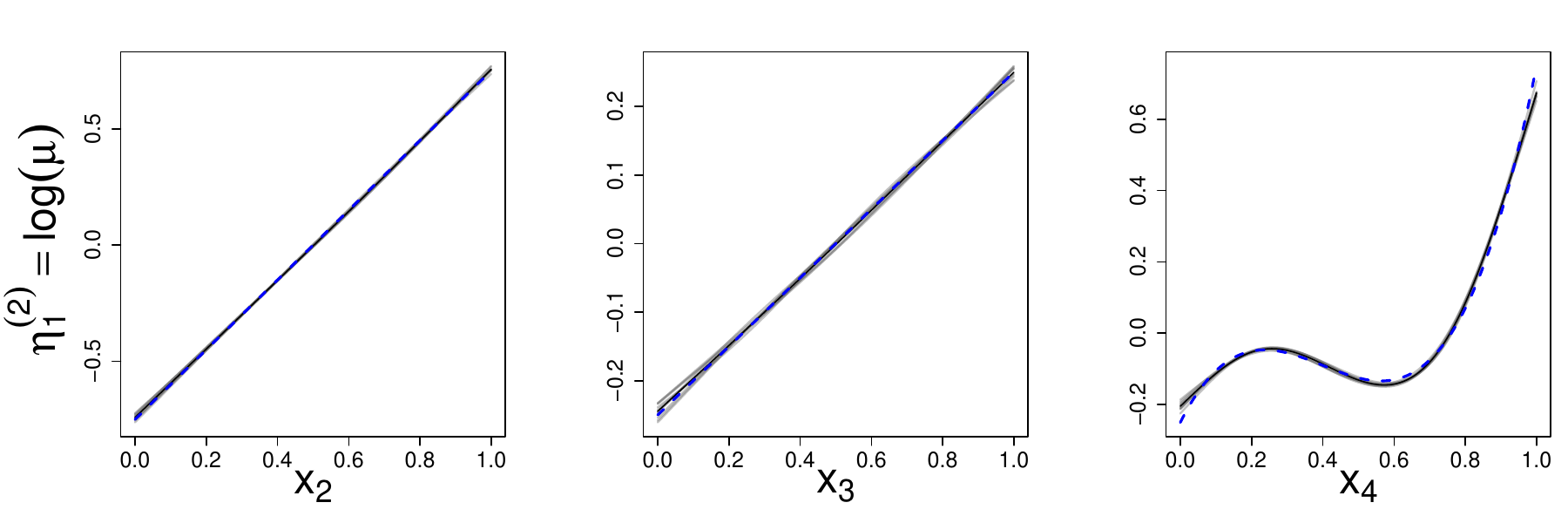}
    \par\medskip
    \makebox[\textwidth][c]{%
        \includegraphics[width=0.50\textwidth]{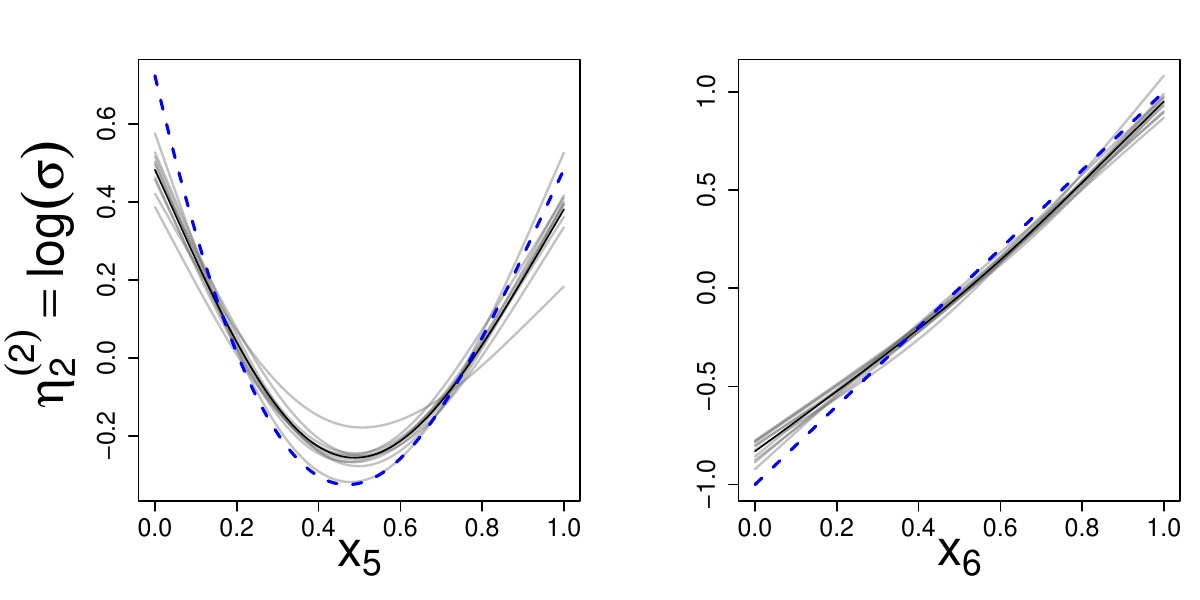}%
        \hspace{0.01\textwidth}%
        \includegraphics[width=0.25\textwidth]{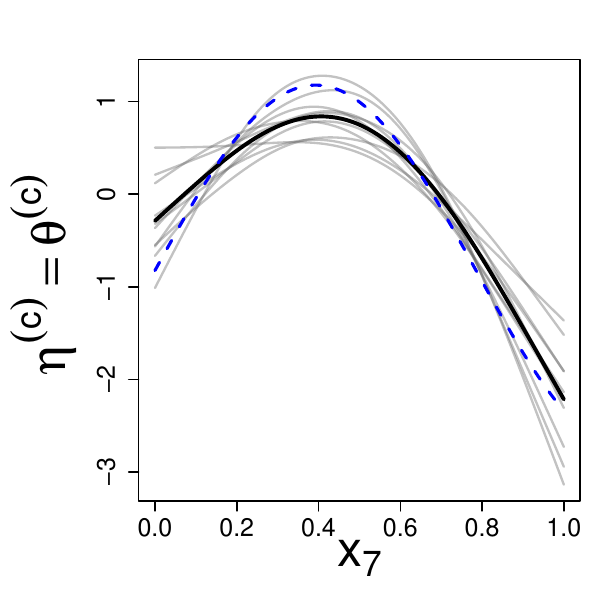}%
        }
    \caption{Partial effects for the seven informative covariates. Blue dashed lines denote the true effects, gray lines the estimates from the 10 replicated datasets, and black solid lines their average.}
    \label{fig:partial-effects}
\end{figure}

\noindent\textbf{Recovery of the covariate effects.} To evaluate how effectively the proposed model retrieves covariate effects for both the distributional and dependence parameters, we analyze the estimated partial effects after fitting. To enhance interpretability, each estimated effect is centered to have mean zero on the link scale so that the intercept corresponds to the overall average prediction. Figure~\ref{fig:partial-effects} shows these centered partial effects on the respective parameter-specific additive predictors, i.e., $\eta_1^{(1)}$, $\eta_1^{(2)}$, $\eta_2^{(2)}$, and $\eta^{(c)}$. The blue dashed curves indicate the true effects, while the solid curves display the estimated effects obtained from 10 replicated datasets. The first row of Figure~\ref{fig:partial-effects} relates to $\eta_1^{(1)}$ in Equation~\ref{eq:simulation-predictors}, which includes a nonlinear contribution of $X_1$ alongside linear effects of $X_2$, $X_3$, and $X_4$. For $X_1$, the estimates reflect the overall upward trend at lower covariate values, but they do not entirely reproduce the true nonlinear form and become clearly flatter at higher values. By comparison, the linear effects of $X_2$, $X_3$, and $X_4$ are recovered with high accuracy across replications. The other rows reveal similarly strong agreement between estimated and true effects for most covariates, covering both linear and nonlinear patterns. For some nonlinear terms, the estimated magnitudes are slightly attenuated relative to the truth. This is in line with the regularization induced by component-wise gradient boosting: small step sizes and early stopping shrink effects toward zero, which reduces estimation variance (\citet{mayr2012generalized}). Taken together, these findings suggest that the proposed model can recover the underlying covariate effects across different distributional parameters while preserving boosting’s regularization. The results therefore demonstrate that the model is intrinsically interpretable at the level of its additive predictors, offering more transparency and auditability than black-box approaches.

\begin{figure}[htbp]
    \centering
    \includegraphics[width=1\textwidth, page=1]{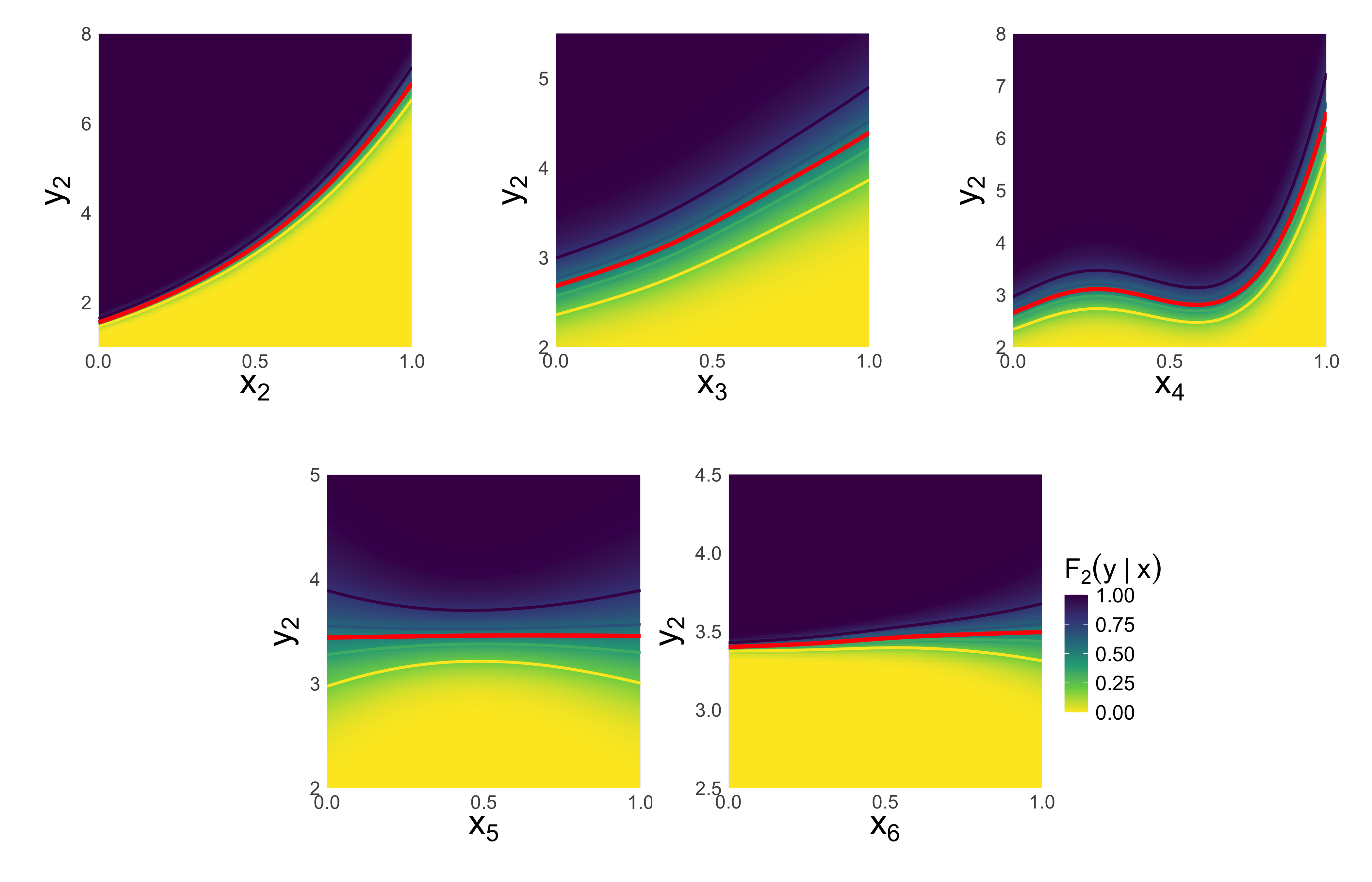}
\caption{Conditional Gamma cumulative distribution functions and quantiles for varying values of the selected covariates, using parameters estimated via the proposed copula-based distributional model.}
\label{fig:conditional-distributions}
\end{figure}
Figure~\ref{fig:conditional-distributions} illustrates the estimated conditional distribution of $Y_2$ and the corresponding conditional quantiles (20\%, 40\%, 60\%, and 80\%) as each covariate $X_j$, $j\in\{2,3,4,5,6\}$, varies. In the simulation setting, $Y_2$ follows a Gamma distribution, with $X_2$, $X_3$, and $X_4$ entering the additive predictor for the conditional mean parameter $\mu$, and $X_5$ and $X_6$ entering the predictor for the scale parameter $\sigma$. The background color represents the estimated conditional CDF, $F_2(y\mid x)$, the red solid line represents the estimated conditional mean, and the colored curves indicate the corresponding conditional quantiles. The resulting distributional patterns closely reflect the covariate effects specified in Equation~\ref{eq:simulation-predictors}. As $X_2$ and $X_3$ increase, the conditional mean and quantiles of $Y_2$ also increase. The larger gaps between the quantile curves indicate an increase in conditional variability. For $X_4$, the conditional mean and quantiles also follow the nonlinear pattern specified in the simulation model. For $X_5$, the conditional spread is relatively large near the lower end of its range, decreases toward intermediate values, and then increases again, closely reflecting the nonlinear effect specified for $\sigma$. Likewise, the increasing effect of $X_6$ on $\sigma$ is reflected in the gradual widening of the conditional distribution as $X_6$ increases. Overall, these results demonstrate that the proposed copula-based distributional model fitted by model-based boosting captures not only changes in the conditional mean but also covariate-dependent changes in the shape and dispersion of the conditional Gamma distribution.

\noindent\textbf{Variable selection performance.} One advantage of model-based boosting is its intrinsic variable selection mechanism, which makes it particularly suitable for high-dimensional settings. At each boosting iteration, only the best-performing base learner is selected and updated, so informative covariates tend to be selected more frequently than non-informative ones.  
\begin{figure}[htbp]
    \centering
    \begin{subfigure}[t]{0.49\textwidth}
        \centering
        \includegraphics[width=\linewidth, page=1]{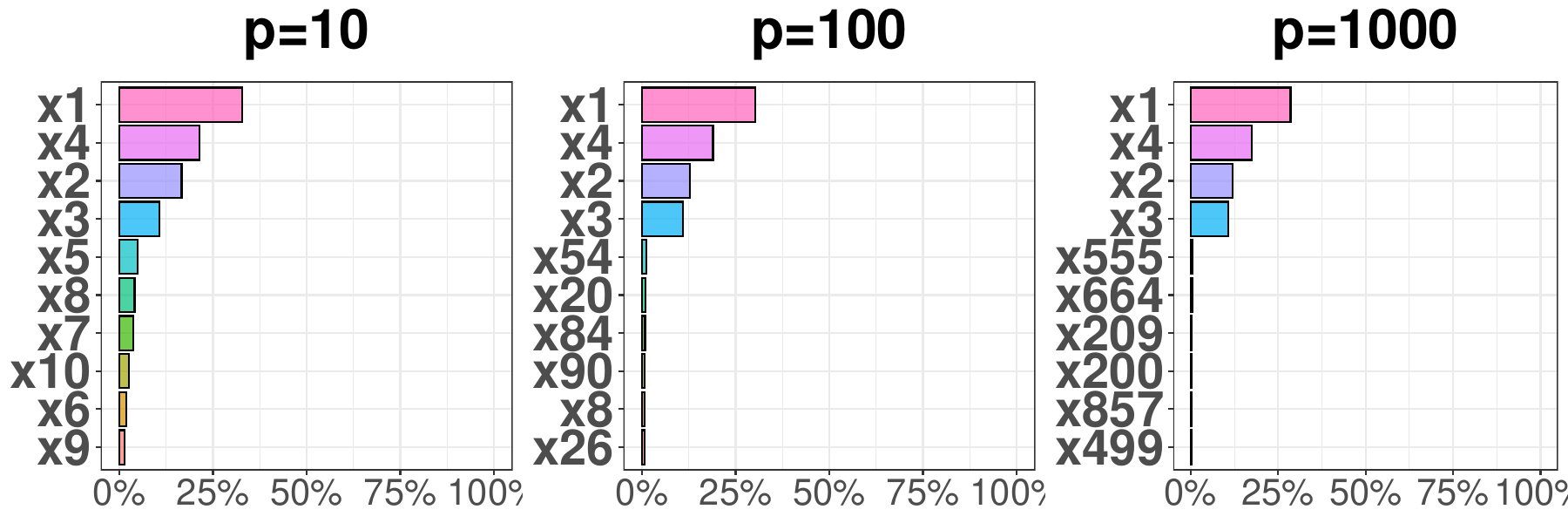}
        \subcaption{}
        \label{fig:var_sel_mu1}
    \end{subfigure}
    \hfill
    \begin{subfigure}[t]{0.49\textwidth}
        \centering
        \includegraphics[width=\linewidth, page=1]{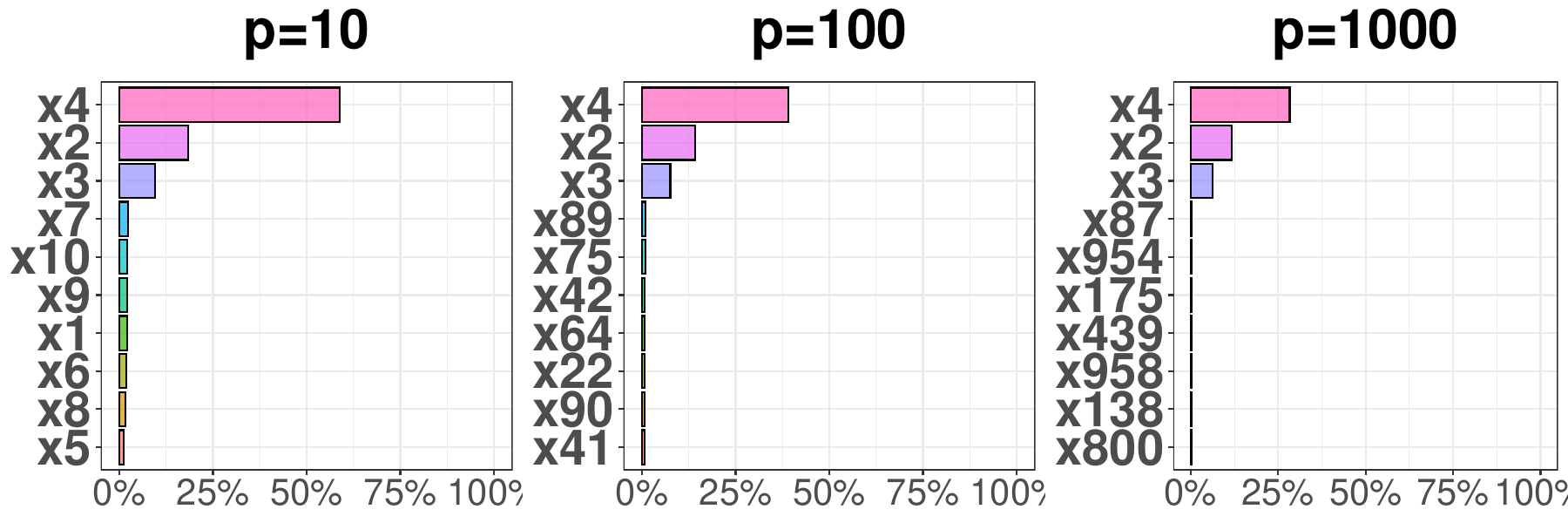}
        \subcaption{}
        \label{fig:var_sel_mu2}
    \end{subfigure}
    \vspace{0.5em}
    \begin{subfigure}[t]{0.49\textwidth}
        \centering
        \includegraphics[width=\linewidth, page=1]{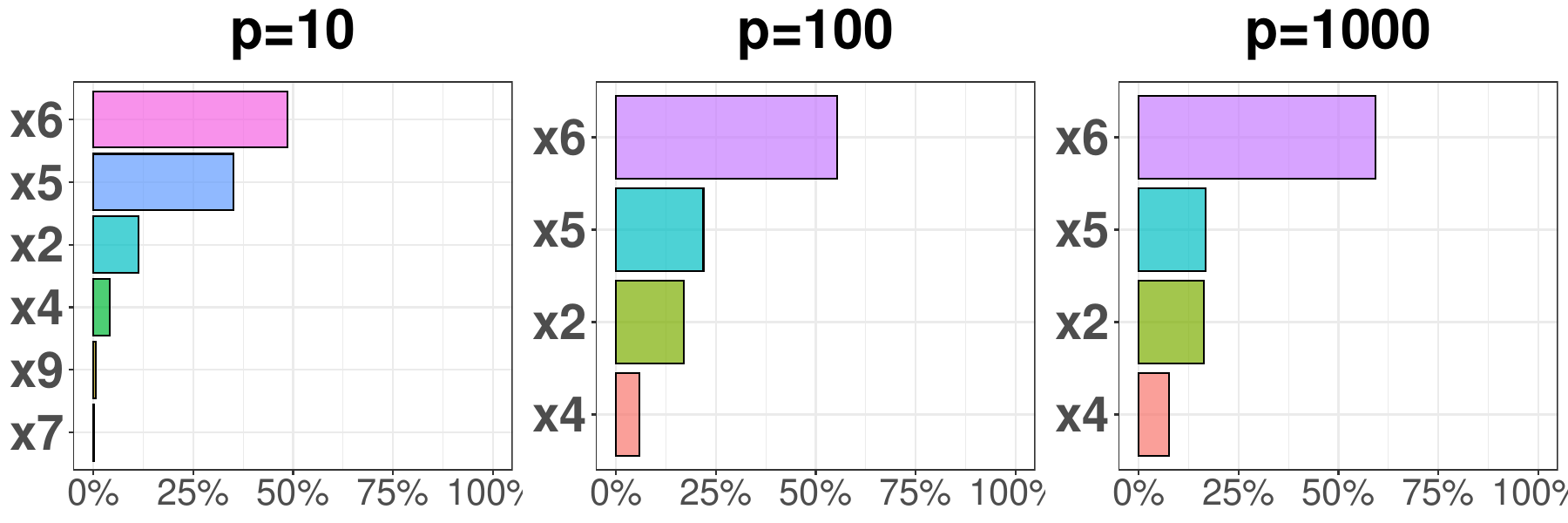}
       \subcaption{}
        \label{fig:var_sel_sigma2}
    \end{subfigure}
    \hfill
    \begin{subfigure}[t]{0.49\textwidth}
        \centering
        \includegraphics[width=\linewidth, page=1]{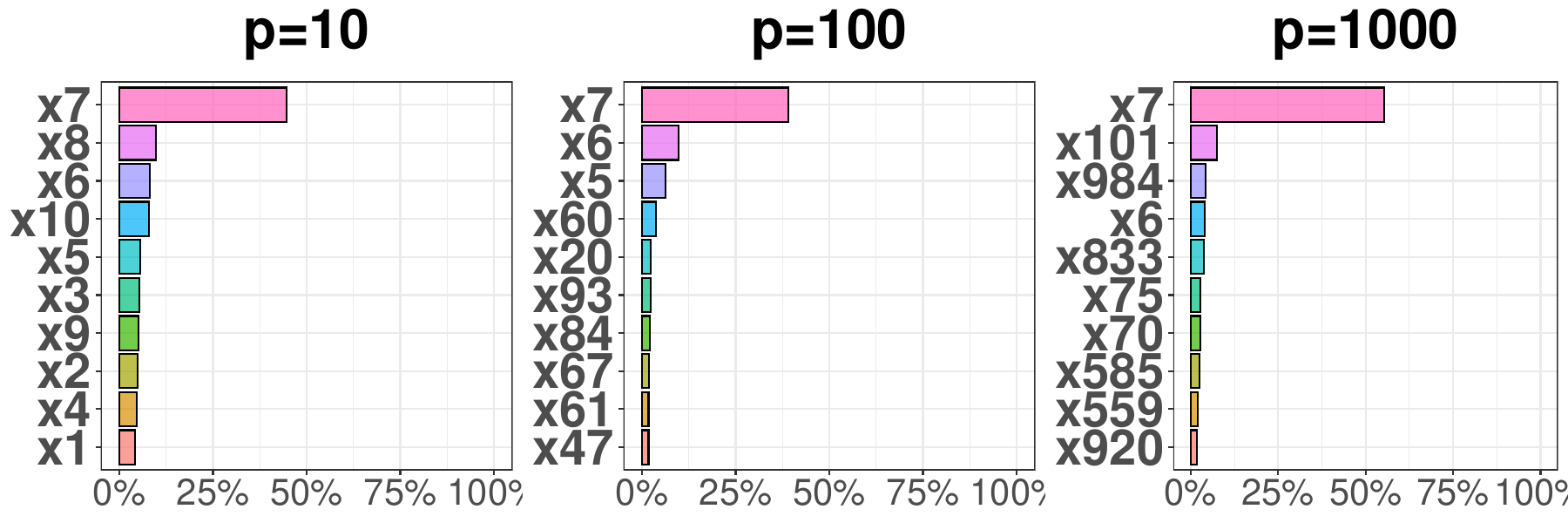}
        \subcaption{}
        \label{fig:var_sel_rho}
    \end{subfigure}
    \caption{Top 10 covariates based on the mean variable selection rate across 10 replications. Panels (a)--(d) correspond to predictors for $\eta_1^{(1)}$, $\eta_1^{(2)}$, $\eta_2^{(2)}$, and $\eta^{(c)}$, respectively.}
    \label{fig:var_sel}
\end{figure}
To examine this behavior, we calculate the selection rate for each covariate and each parameter-specific predictor. We then average these rates over the 10 replications, and Figure~\ref{fig:var_sel} reports the 10 covariates with the highest mean selection rates for $p=10$, $100$, and $1000$. In Figure~\ref{fig:var_sel_sigma2}, fewer than 10 covariates are displayed because the remaining covariates were never selected in any of the 10 replications, resulting in mean selection rates of zero. In all three scenarios, the covariates used to construct each parameter-specific predictor consistently appear among the most frequently selected variables. For $\eta_1^{(1)}$, the truly informative covariates $X_1$, $X_2$, $X_3$, and $X_4$ take the top positions. Likewise, $X_2$, $X_3$, and $X_4$ are selected most often for $\eta_1^{(2)}$, whereas selections for $\eta_2^{(2)}$ are dominated by $X_5$ and $X_6$. For the dependence predictor $\eta^{(c)}$, $X_7$ is consistently the most frequently chosen covariate. Notably, these patterns remain evident as the dimension grows from $p=10$ to $p=1000$, including the case where the number of covariates matches the number of training observations ($p=n=1000$). While some non-informative covariates enter the top 10, their selection rates stay markedly below those of the genuinely informative covariates. Overall, these findings indicate that the proposed boosting procedure reliably recovers informative covariates while curbing the impact of irrelevant variables, even in high-dimensional settings.

\section{Empirical Investigation} \label{sec:sc5}
We examine the proposed distributional copula regression approach, estimated via model-based boosting, employing Poisson and Gamma marginals for claim frequency and severity, respectively, using car insurance claims data that include telematics information.

\subsection{Data and Model Specification}
\label{subsec:empirical_setup}
To evaluate the empirical performance of the proposed framework, we used the synthetic telematics dataset developed by \citet{so2021synthetic}, which is publicly available at \url{http://www2.math.uconn.edu/~valdez/data.html}. The dataset was designed to emulate key characteristics of automobile telematics data obtained from a Canadian insurer, providing a publicly accessible benchmark without requiring access to the original confidential policyholder records. It contains 100,000 observations and 52 variables, covering traditional insurance characteristics, telematics measurements, and claims outcomes. Variable descriptions are provided in Table \ref{tab:VD2} in Appendix C.

Prior to model fitting, we restructured or excluded potentially redundant covariates, guided by the correlation analysis reported in \citet{so2025interpretable}. The preprocessing focused primarily on groups with correlation coefficients exceeding approximately 0.80 in magnitude. The three nested driving-duration proportions were converted into covariates representing driving durations between two and three hours, between three and four hours, and beyond four hours. These were obtained by subtracting \texttt{Pct.drive.3hrs} from \texttt{Pct.drive.2hrs}, subtracting \texttt{Pct.drive.4hrs} from \texttt{Pct.drive.3hrs}, and retaining \texttt{Pct.drive.4hrs}, respectively. Acceleration, braking, and left- and right-turn measurements were summarized separately using logarithmically transformed weighted sums. Weights of 1--6 were assigned to the acceleration and braking thresholds, and weights of 1--5 to the turning-intensity thresholds, in ascending order. This construction places greater emphasis on measurements associated with more extreme driving behavior. The weekday and weekend proportions, \texttt{Pct.drive.wkday} and \texttt{Pct.drive.wkend}, were excluded because they summarize driving activity already represented by the day-of-week proportions. Six day-of-week proportions, corresponding to Monday through Saturday, were retained as candidate predictors, with Sunday omitted as the remaining component. We also excluded \texttt{Annual.pct.driven} because its correlation of approximately 0.70 with \texttt{Total.miles.driven} indicated overlapping information. Expected annual mileage (\texttt{Annual.miles.drive}) and total mileage were transformed using \(\log(1+x)\) to compress large values and reduce right-skewness. 

The responses were claim count, \texttt{NB\_Claim}, and average claim amount. For policies with positive claim counts, the average claim amount was calculated as \texttt{AMT\_Claim} divided by \texttt{NB\_Claim}. The resulting candidate predictor set comprised five categorical covariates: insured sex, marital status, vehicle use, region, and territory, and 22 numerical covariates. The distributional copula regression model, with Poisson and Gamma marginals for claim count and average claim amount, respectively, was fitted using non-cyclic model-based boosting. For each distributional parameter, the additive predictor was specified using the same candidate base-learner set, with covariate effects estimated separately for each parameter. Categorical covariates were dummy coded and represented by ordinary linear base-learners. Numerical covariates were represented by B-spline base learners with degrees of freedom set to three, allowing both linear and smooth nonlinear associations. The claim frequency predictor \(\eta_{1}^{(1)}\) additionally included the logarithm of policy exposure as an offset, with its coefficient fixed at one.

For model evaluation, the observations were randomly divided into training and test sets in an 80:20 ratio. The training set was further divided into fitting and validation subsets in an 80:20 ratio. The step length for each parameter predictor was selected by grid search over \{0.001, 0.005, 0.01, 0.05, 0.1\}, and the maximum number of boosting iterations was set to 2,000. Distributional copula regression models with Frank, Gumbel, and Gaussian copulas were fitted and compared with the two benchmark models described in Section~\ref{sec:sc4}. All models used the same candidate base-learner set to ensure a consistent comparison. Data analysis and model estimation were implemented in R and Python.

\subsection{Empirical Results}
\label{subsec:empirical_results}
\begin{table}[htbp]
\centering
\caption{Predictive performance on the synthetic telematics dataset.}
\label{tab:Empirical_performance_summary}
\small
\renewcommand{\arraystretch}{1.2}
\begin{tabular}{lcc}
\toprule
\textbf{Model}
& \textbf{Negative log-likelihood}
& \textbf{Energy score} \\
\midrule
Frank copula boosting
    & \color{red} -305,276 & \color{red}130.8 \\
Gaussian copula boosting
    & -305,228 & 131.0 \\
Gumbel copula boosting
    & -305,065 & 131.3 \\
Univariate boosting
    & -302,762 & 134.4 \\
Univariate MLE
    & -297,023 & 132.6 \\
\bottomrule
\end{tabular}
\end{table}
\noindent Table~\ref{tab:Empirical_performance_summary} indicates that the three copula boosting models yield smaller negative log-likelihood and energy score values than the two univariate benchmark models. Within the copula family, the Frank copula delivers the strongest performance on both metrics, with the Gaussian copula next and the Gumbel copula last. Overall, these findings imply that explicitly capturing the conditional dependence between claim frequency and mean claim severity enhances joint predictive accuracy, while also providing important evidence about the form of dependence between frequency and severity. In particular, the estimated Kendall's $\tau$ for the Frank copula varies substantially, ranging from $-0.157$ to $0.72$, which suggests that both the sign and strength of dependence differ markedly across policyholder risk profiles. The occurrence of negative dependence ($\tau < 0$) points to subgroups of policyholders who tend to file claims more often but with lower average costs. Since the standard Gumbel copula is limited to positive dependence and enforces pronounced upper-tail dependence, its weaker results indicate that the relationship between claim frequency and severity is not purely positive. This also suggests that very high claim counts and extremely severe losses do not necessarily coincide in the tail.

\begin{figure}[htbp]
    \centering
    \begin{subfigure}[t]{0.24\textwidth}
        \centering
        \includegraphics[width=\linewidth, page=1]{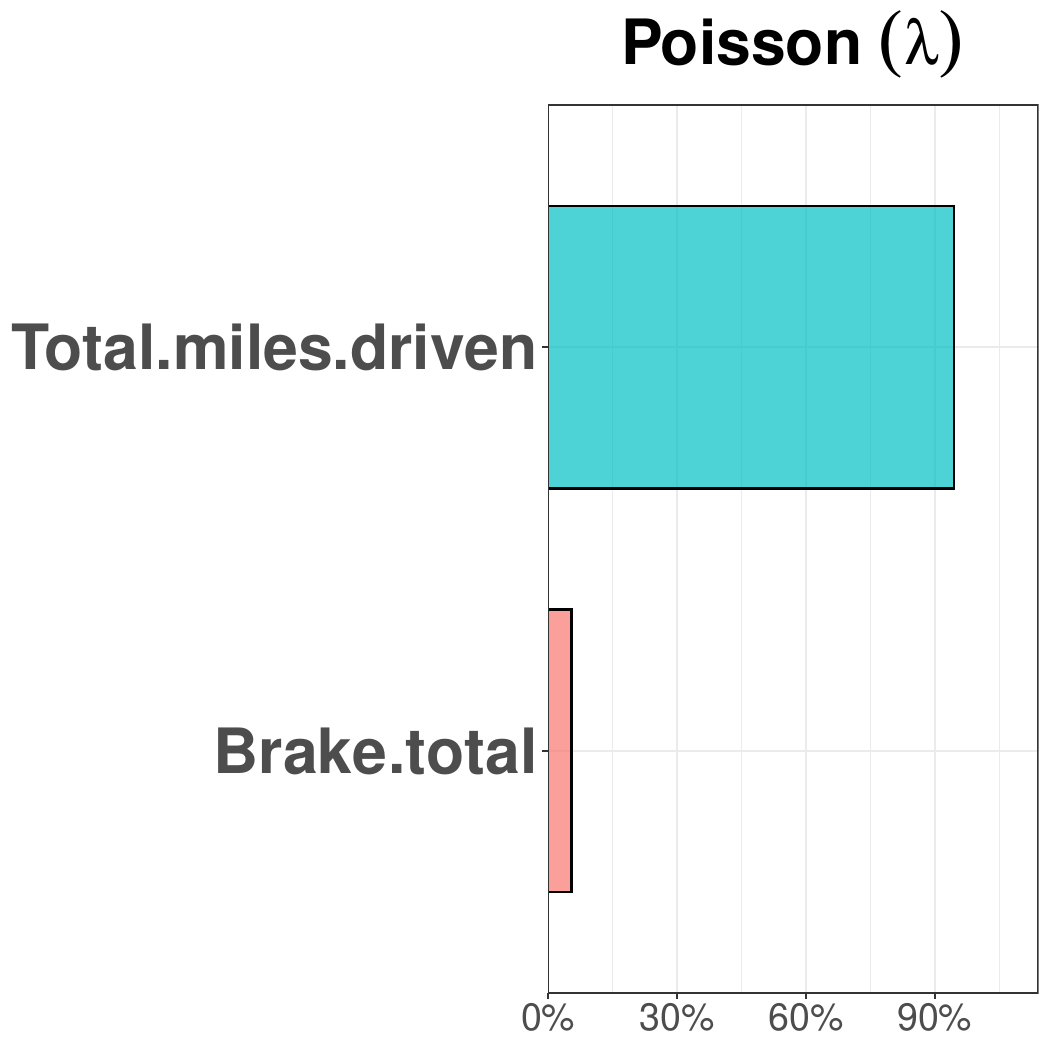}
        \subcaption{}
        \label{fig:var_sel_mu1_real}
    \end{subfigure}
    \hfill
    \begin{subfigure}[t]{0.24\textwidth}
        \centering
        \includegraphics[width=\linewidth, page=1]{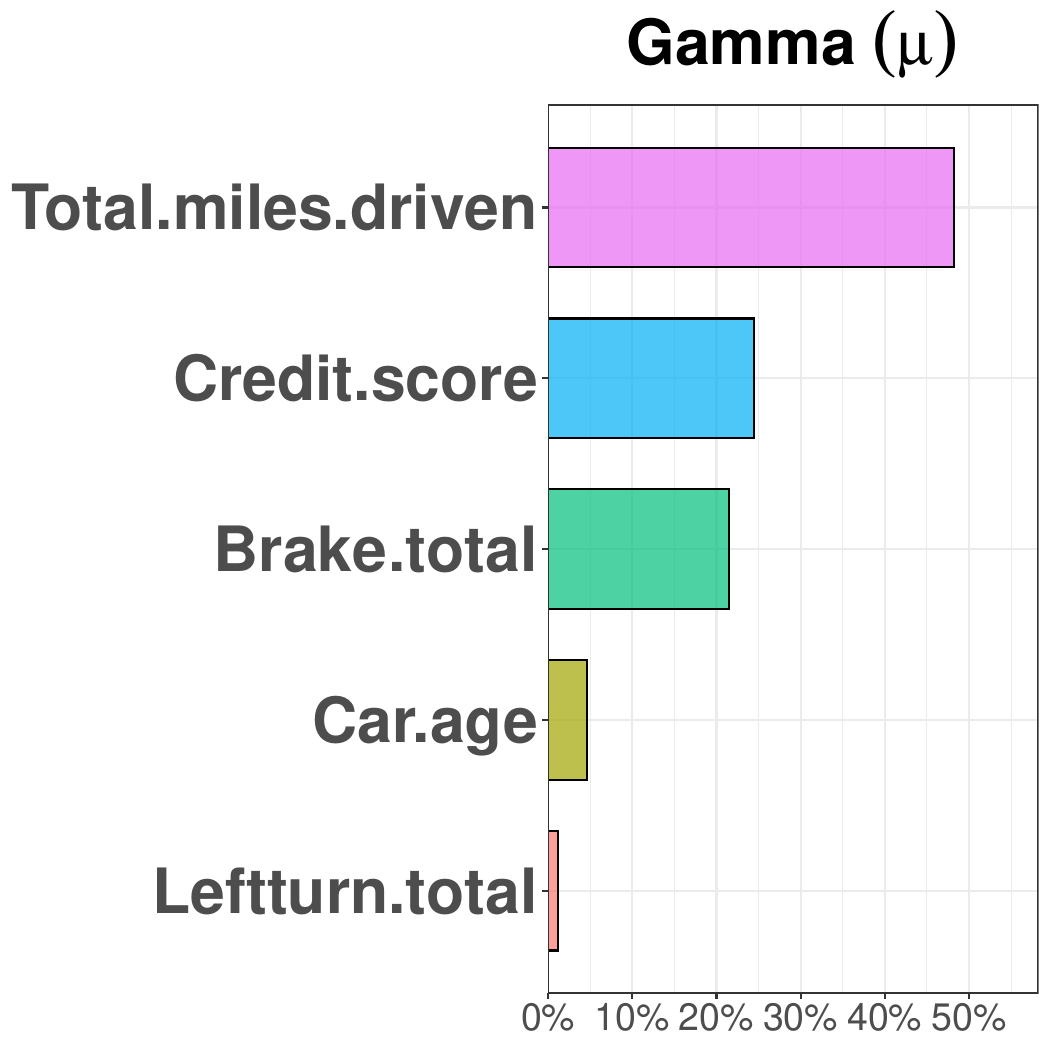}
        \subcaption{}
        \label{fig:var_sel_mu2_real}
    \end{subfigure}
    \vspace{0.5em}
    \begin{subfigure}[t]{0.24\textwidth}
        \centering
        \includegraphics[width=\linewidth, page=1]{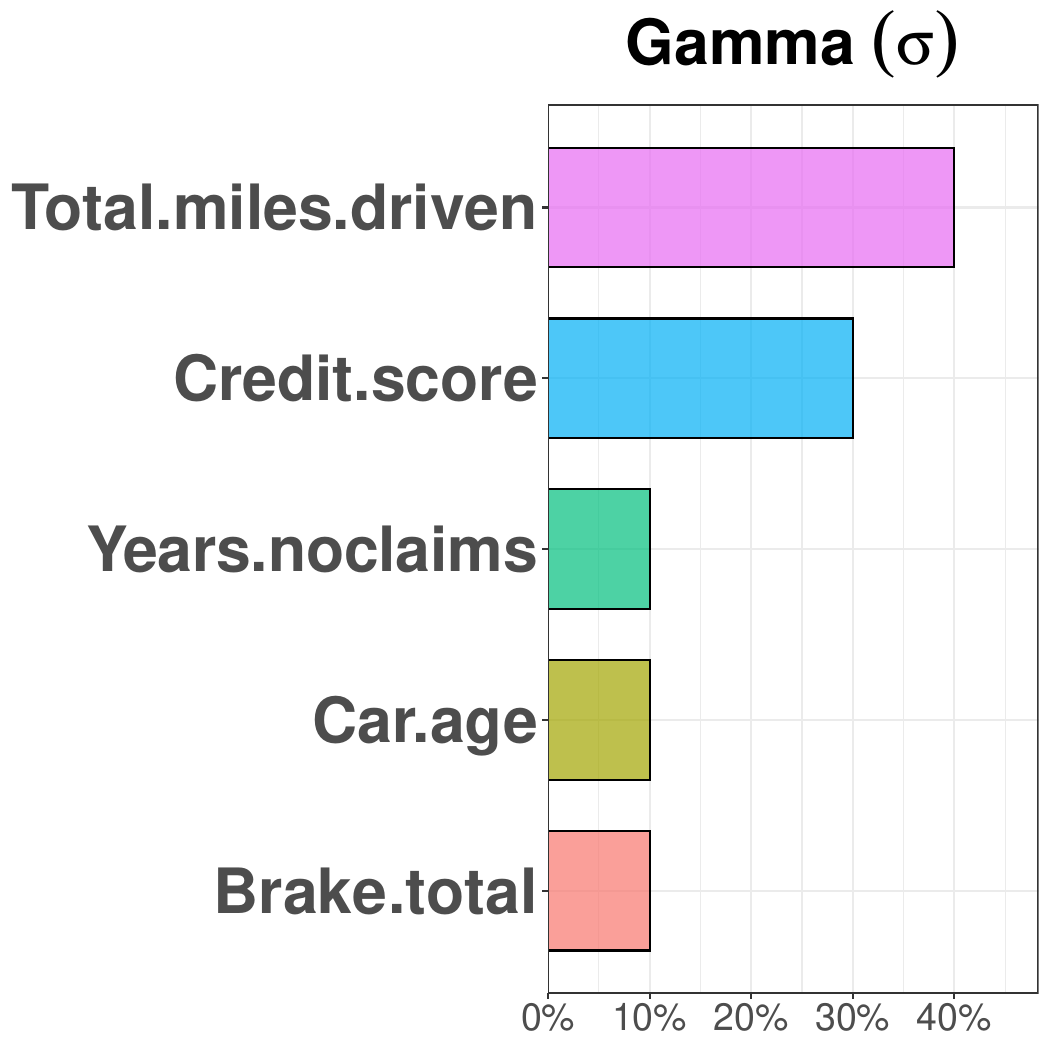}
       \subcaption{}
        \label{fig:var_sel_sigma2_real}
    \end{subfigure}
    \hfill
    \begin{subfigure}[t]{0.24\textwidth}
        \centering
        \includegraphics[width=\linewidth, page=1]{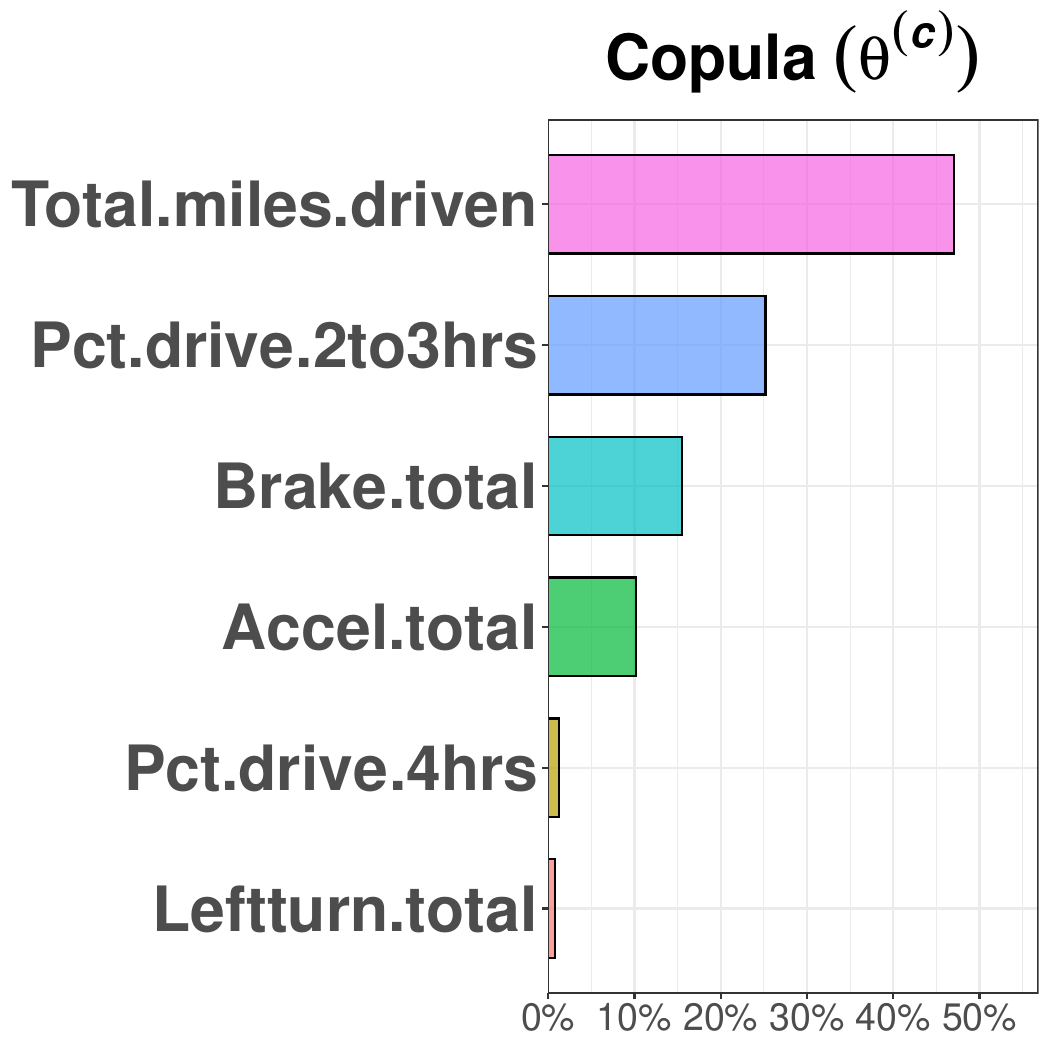}
        \subcaption{}
        \label{fig:var_sel_rho_real}
    \end{subfigure}
    \caption{Covariates included in the predictor for each parameter, ranked by selection rate.}
    \label{fig:var_sel_real}
\end{figure}

Given its superior predictive performance, the subsequent analyses focus on the fitted Frank copula boosting model. 
Figure~\ref{fig:var_sel_real}, panels (a)--(d), shows the covariates selected for the additive predictors \(\eta_1^{(1)}\), \(\eta_1^{(2)}\), \(\eta_2^{(2)}\), and \(\eta^{(c)}\), respectively, ordered by their boosting selection rates. In the claim frequency component, the predictor for $\lambda$ retained only two covariates—\texttt{Total.miles.driven} and \texttt{Brake.total}—as informative variables, in addition to the fixed exposure offset. \texttt{Total.miles.driven} contributed to over 90\% of the updates, suggesting that most revisions to this predictor were driven by its effect. For claim severity, the $\mu$ predictor had three covariates with selection rates above 20\%: \texttt{Total.miles.driven}, \texttt{Credit.score}, and \texttt{Brake.total}. In the dispersion predictor for severity, only \texttt{Total.miles.driven} and \texttt{Credit.score} surpassed this cutoff, whereas \texttt{Years.noclaims}, \texttt{Car.age}, and \texttt{Brake.total} each showed selection rates around 10\%. The total miles driven variable was repeatedly chosen in both the frequency and severity models, which aligns with the intuition that higher driving exposure raises both accident likelihood and the potential magnitude of losses. Consistent patterns were documented by \citet{so2025zero}, who examined the same dataset and reported associations between claim amounts and total mileage, credit score, vehicle age, and braking and turning behavior, among others. For the copula dependence predictor, the four most frequently selected covariates were \texttt{Total.miles.driven}, \texttt{Pct.drive.2to3hrs}, \texttt{Brake.total}, and \texttt{Accel.total}, though only the first two achieved selection rates exceeding 20\%. Overall, \texttt{Total.miles.driven} emerged as informative across all four predictors and was selected most often, highlighting its key role in modeling claim frequency, claim severity, severity dispersion, and their conditional dependence.

\begin{figure}[htbp]
    \centering
    \begin{subfigure}[t]{0.24\textwidth}
        \centering
        \includegraphics[width=\linewidth, page=1]{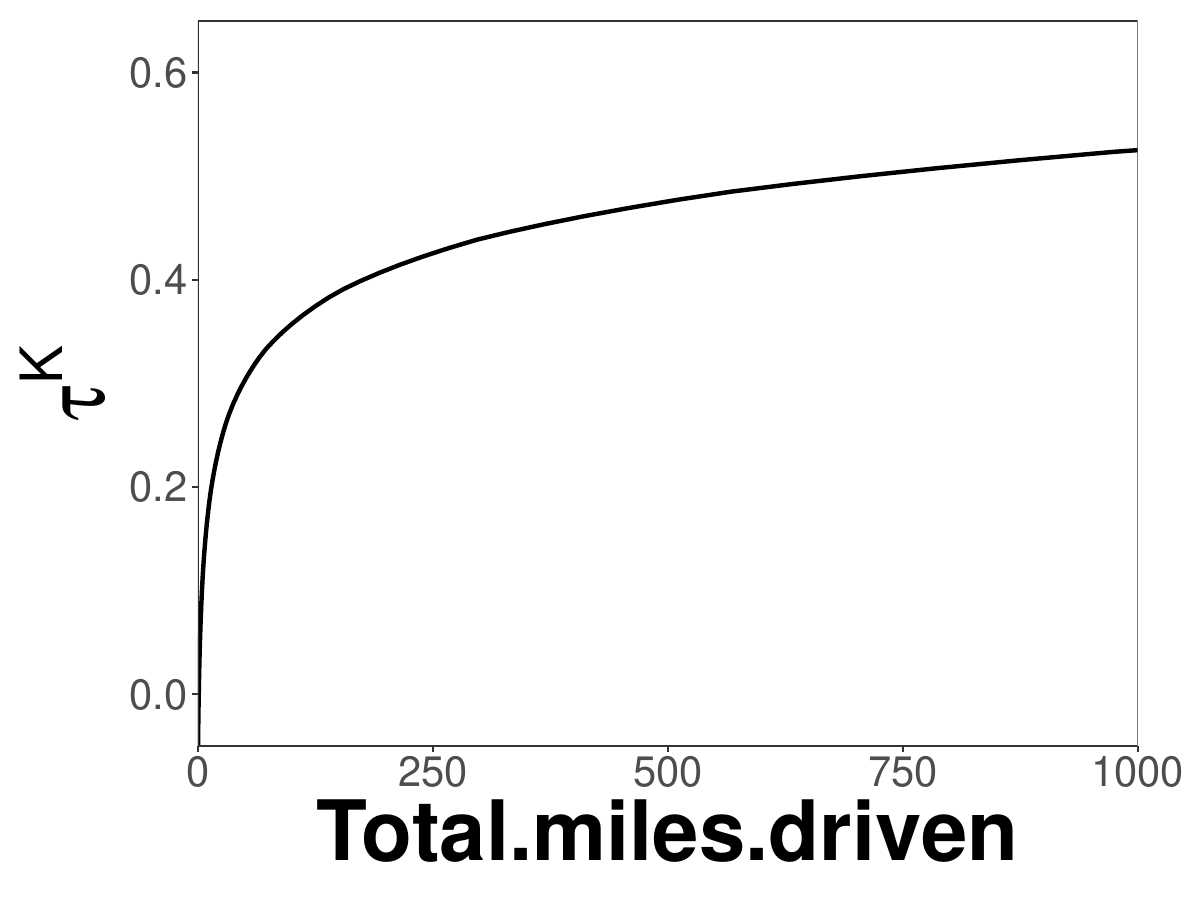}
        \subcaption{}
        \label{fig:Tau1_real}
    \end{subfigure}
    \hfill
    \begin{subfigure}[t]{0.24\textwidth}
        \centering
        \includegraphics[width=\linewidth, page=1]{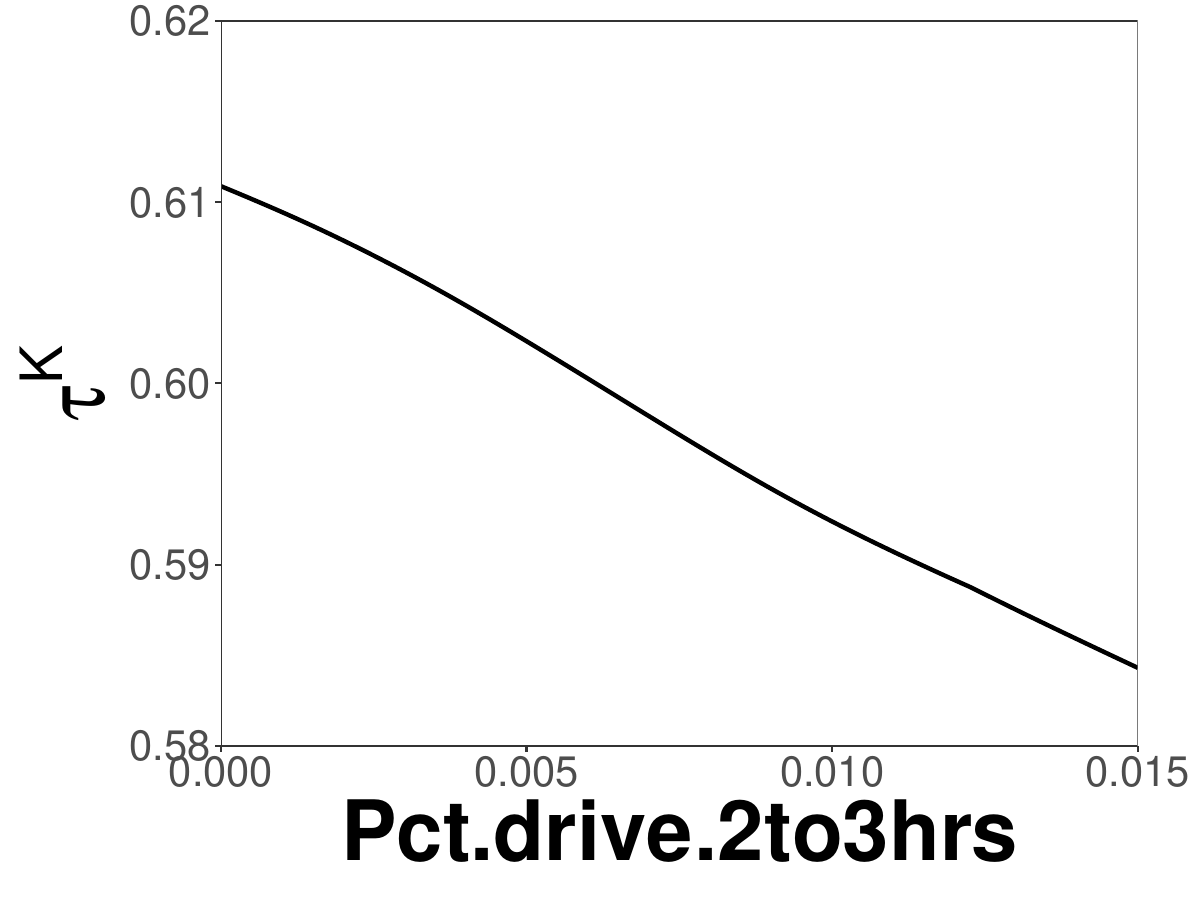}
        \subcaption{}
        \label{fig:Tau2_real}
    \end{subfigure}
    \vspace{0.5em}
    \begin{subfigure}[t]{0.24\textwidth}
        \centering
        \includegraphics[width=\linewidth, page=1]{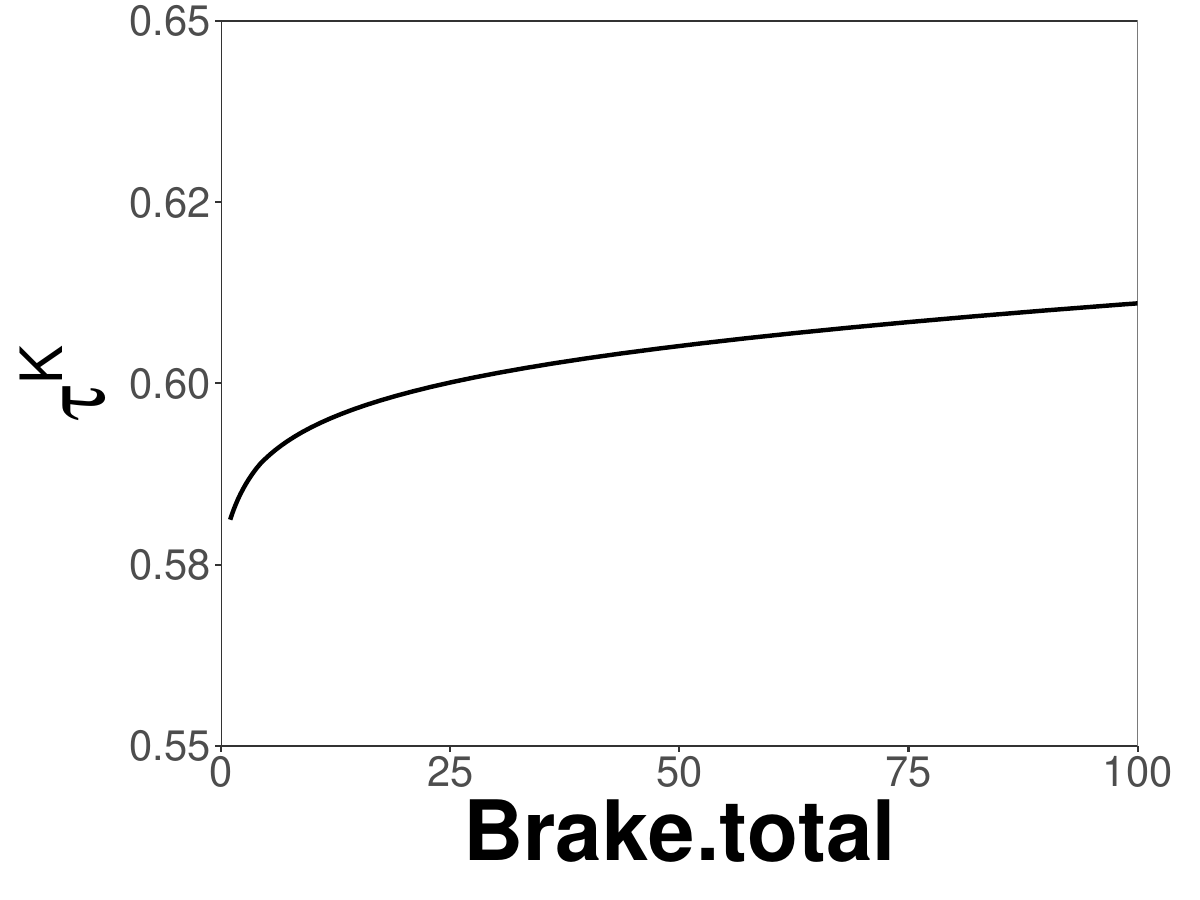}
       \subcaption{}
        \label{fig:Tau3_real}
    \end{subfigure}
    \hfill
    \begin{subfigure}[t]{0.24\textwidth}
        \centering
        \includegraphics[width=\linewidth, page=1]{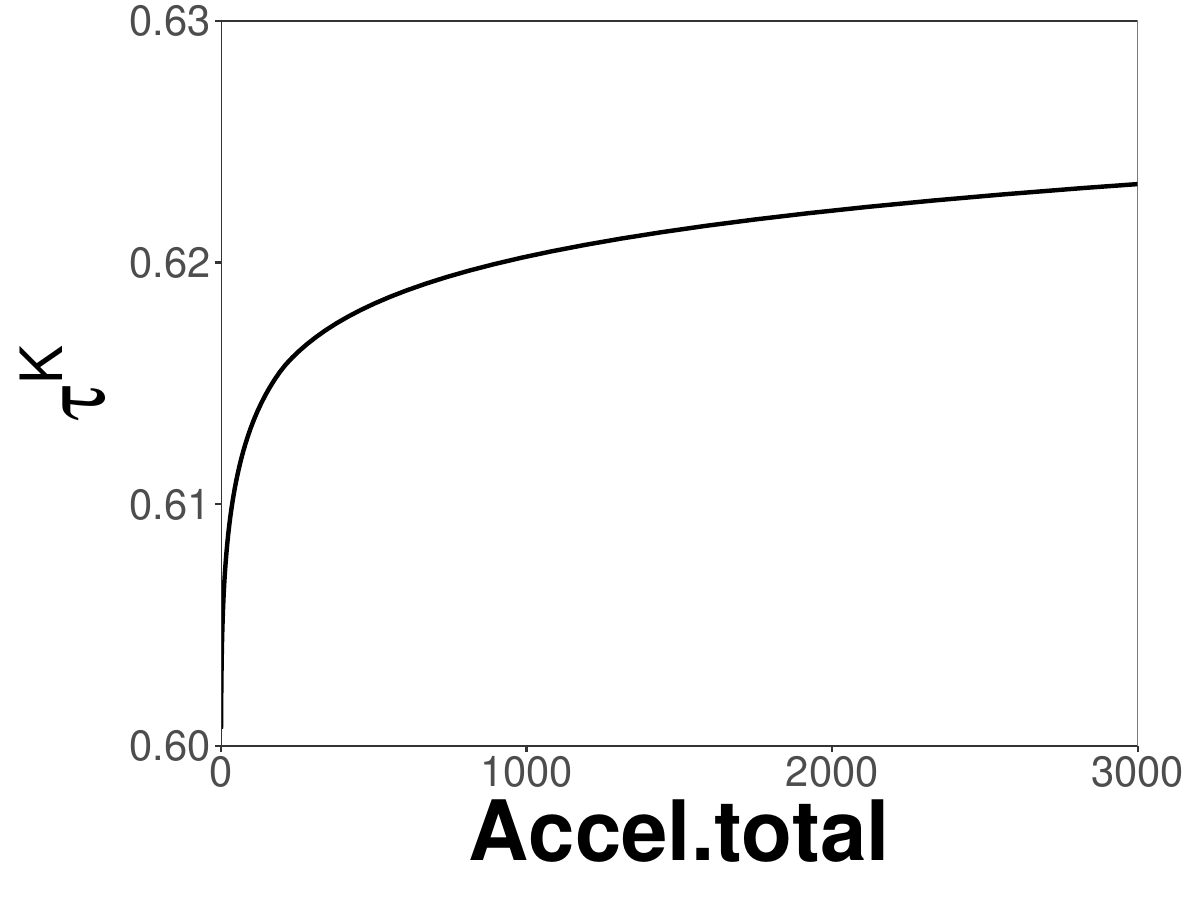}
        \subcaption{}
        \label{fig:Tau4_real}
    \end{subfigure}
    \caption{Effects of covariates on Kendall's tau rank correlation coefficient between claim frequency and severity}
    \label{fig:Tau_real}
\end{figure}

We next turn to the four covariates that are most frequently selected for the Frank copula dependence predictor \(\eta^{(c)}\). Figure~\ref{fig:Tau_real} illustrates how the model-implied Kendall’s $\tau$ varies with these four covariates. For Total.miles.driven (Figure~\ref{fig:Tau1_real}), the relationship is close to zero—or mildly negative—at very low distances, then rises rapidly before flattening. This pattern is consistent with actuarial reasoning: policyholders with minimal driving exposure typically experience minor, low-severity incidents, resulting in weak or negative dependence between frequency and severity. As total mileage grows, exposure expands to riskier settings (e.g., highways and complicated intersections), producing a pronounced positive association in which higher claim counts are much more likely to coincide with higher severities. The proportion of continuous driving between 2 and 3 hours (Figure~\ref{fig:Tau2_real}), used as a surrogate for fatigue and long-distance travel, shows a modest declining trend while remaining at a consistently high dependence level over its full domain ($\tau \approx 0.58 \text{ to } 0.61$). This sustained magnitude of $\tau$ highlights the substantial tail risk of extended driving, where crashes are intrinsically more prone to be catastrophic. The gentle negative slope may indicate that drivers with very high proportions of such trips are seasoned long-distance commuters or professional drivers; their experience with uninterrupted highway travel slightly reduces the extreme joint risk linking frequency and severity, even though dependence remains strong. Brake.total and Accel.total are both key markers of aggressive driving. Figure~\ref{fig:Tau_real} shows a sharp early increase in $\tau$, followed by saturation at larger numbers of events. Thus, as harsh braking and rapid acceleration become more common, the probability of observing both frequent and severe claims together increases markedly—aggressive drivers tend to crash more often and with greater collision forces. Taken together, these non-linear, covariate-dependent shifts in Kendall’s $\tau$ reinforce the importance of modeling the dependence structure in a dynamic way.

\begin{figure}[htbp]
    \centering      \includegraphics[width=0.6\textwidth]{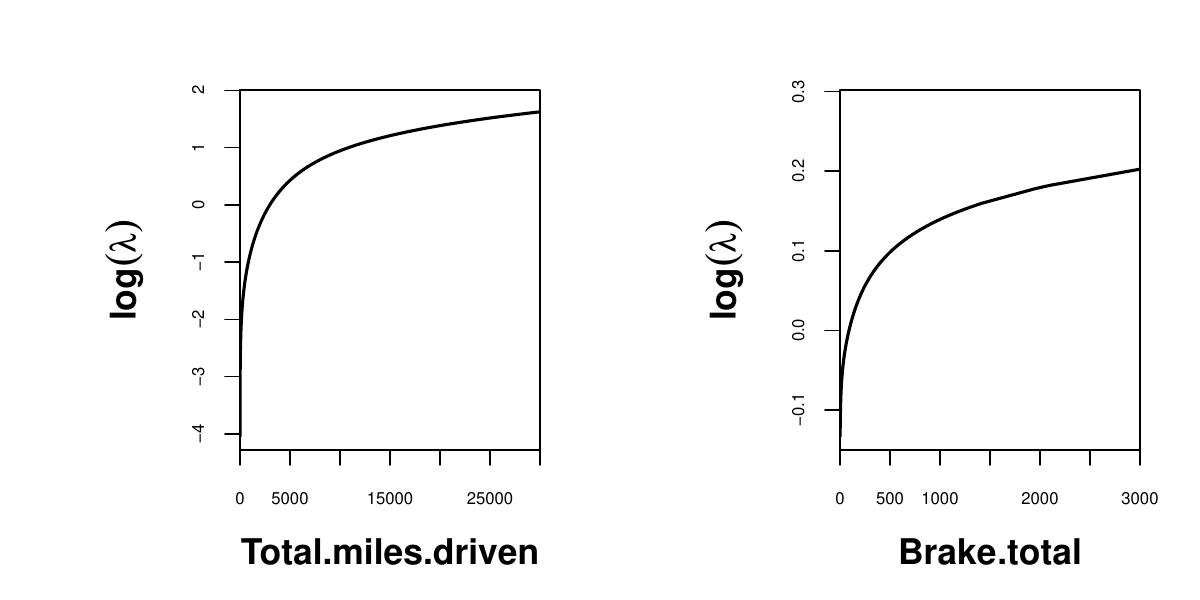}
    \caption{Estimated partial effects on the additive predictor for the Poisson mean}
    \label{fig:poisson_effects_real}
\end{figure}

In Figure~\ref{fig:poisson_effects_real}, the estimated partial effects show that both total distance driven and the braking index are positively associated with expected claim counts. The effect of total mileage increases sharply at low mileage levels and more gradually thereafter on the log scale. This pattern is consistent with greater driving activity providing more opportunities for claims to occur. The braking index also exhibits an increasing effect, suggesting that more frequent or more intense braking is associated with higher claim frequency.

\begin{figure}[htbp]
    \centering
        \includegraphics[width=1\textwidth]{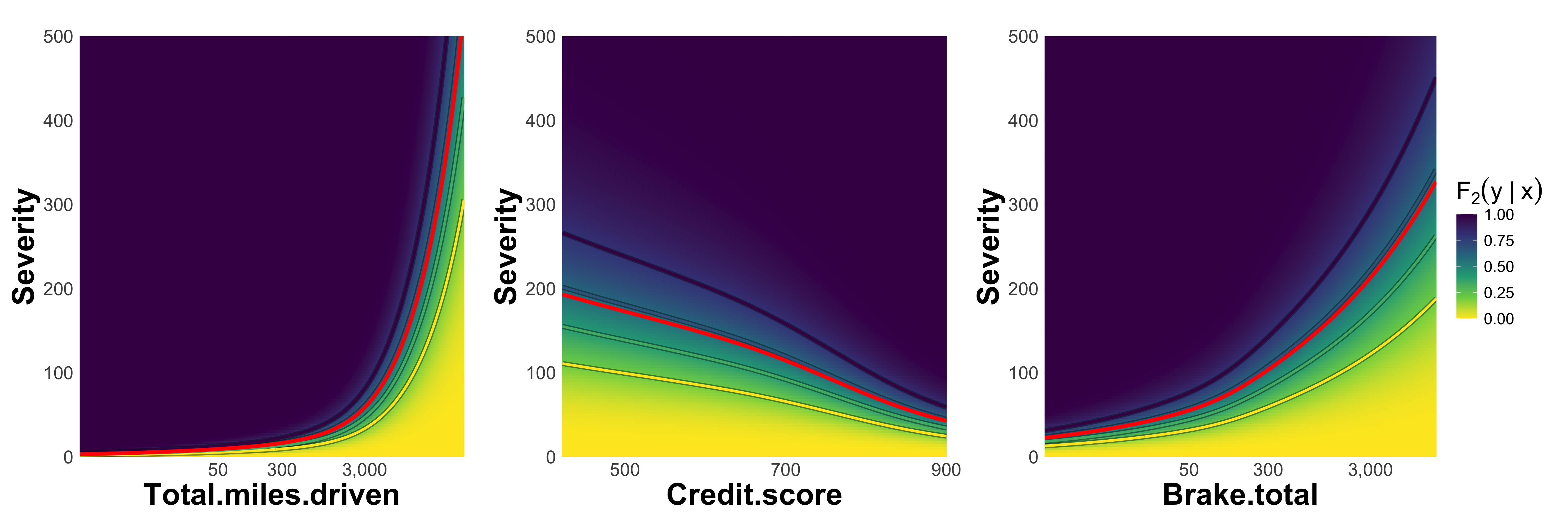}
    \caption{Conditional Gamma cumulative distributions and quantiles of claim severity for varying values of the informative covariates.}
    \label{fig:CDF_real}
\end{figure}
To examine how covariates affect claim severity, we graphed the estimated conditional Gamma cumulative distribution functions and quantiles over a range of covariate values. Since the model simultaneously fits the mean and the dispersion of severity, these visualizations show how covariates modify the entire fitted distribution rather than only its center. Figure~\ref{fig:CDF_real} displays results for \texttt{Total.miles.driven}, \texttt{Credit.score}, and \texttt{Brake.total}, which were chosen due to their strong contributions to the mean and dispersion components (see Figures~\ref{fig:var_sel_mu2_real} and~\ref{fig:var_sel_sigma2_real}). The red solid curve denotes the estimated conditional mean claim amount. For \texttt{Total.miles.driven}, the severity quantiles rise steeply, and the distance between lower and upper quantiles expands markedly. This indicates that higher driving exposure increases not only expected severity but also variability and tail risk in claim amounts. By comparison, higher \texttt{Credit.score} shifts the quantiles downward and compresses their spread. From an actuarial perspective, credit-based insurance scores are widely viewed as a robust proxy for risk aversion and responsible conduct. The tighter distribution at high credit scores suggests these insureds tend to generate more predictable, low-severity claims. The \texttt{Brake.total} measure also trends upward, with upper quantiles increasing faster than lower quantiles. As a telematics-based signal of aggressive driving, frequent hard braking can reflect tailgating or slower reaction times. The resulting asymmetric widening suggests that aggressive driving disproportionately increases the right tail of the severity distribution, substantially raising the likelihood of extreme, high-cost claims. Overall, these findings highlight the usefulness of distributional regression for actuarial pricing.



In conclusion, results from the synthetic telematics dataset show that the proposed framework achieves higher predictive accuracy while maintaining interpretability. The boosting procedure selects a smaller subset of candidate covariates for each distributional parameter, yielding more parsimonious predictor specifications. Since the approach models the full conditional distribution via a copula, it also makes explicit how covariates relate to both the level and variability of claim severity, as well as to the dependence between claim frequency and severity. Taken together, these features support transparent model assessment and highlight the framework’s potential in insurance applications that demand both strong predictive performance and clear interpretability.

\section{Concluding Remarks} \label{sec:conclude}
This paper presented a distributional copula regression framework for jointly modeling claim frequency and claim severity. It integrates interpretable, parameter-specific additive predictors for the marginal distributions with a copula whose dependence can vary with covariates, resulting in an inherently interpretable (glass-box) model. Estimation is carried out via noncyclic model-based boosting, producing sparse additive predictors and enabling automatic variable selection separately for each marginal parameter as well as for the copula dependence parameter. A simulation study and an empirical analysis on a synthetic telematics insurance dataset show that the proposed approach can enhance out-of-sample predictive performance compared with univariate benchmarks while maintaining interpretability. Partial effect plots and conditional distributional summaries demonstrate how covariates influence specific distributional parameters and the implied predictive distributions. Notably, the covariate-dependent copula offers a clear explanation of how driving-related characteristics affect the conditional association between claim frequency and severity. By uniting predictive performance, variable selection, dependence modeling between frequency and severity, and interpretable distributional regression, the framework provides a promising tool for insurance settings that require both accurate forecasts and transparent model explanations.

\clearpage
\subsection*{Appendix A. Poisson--Gamma responses with a Gaussian copula}
\label{appb-application}
We consider Poisson--Gamma marginal distributions linked through a Gaussian copula. We assume that the first marginal response follows a Poisson distribution,
\[Y_1\sim\operatorname{Poisson}(\lambda),
\qquad \lambda>0,\]
and that the second marginal response follows a Gamma distribution,
\[Y_2\sim\operatorname{Gamma}(\mu,\sigma),
\qquad \mu>0,\quad \sigma>0,\]
where \(\mu\) is the mean severity and \(\sigma\) is a dispersion parameter.

For the Poisson margin, the probability mass function and CDF are
\begin{equation}
\label{eq:poisson-margin}
f_1=f_1(y_1\mid\lambda)
=\frac{\exp(-\lambda)\lambda^{y_1}}{y_1!}, \qquad
F_1=F_1(y_1\mid\lambda)=\sum_{j=0}^{y_1}\frac{\lambda^j\exp(-\lambda)}{j!},
\end{equation}
where \(y_1=0,1,2,\ldots\). Because \(Y_1\) is discrete, we also define the left limit of its CDF as
\begin{equation}
\label{eq:poisson-left-limit}
F_1^{-}=\Pr(Y_1<y_1)=\sum_{j=0}^{y_1-1}\frac{\lambda^j\exp(-\lambda)}{j!},
\end{equation}
with the convention that \(F_1^{-}=0\) when \(y_1=0\).

Under the mean--dispersion parameterization of the Gamma distribution,
\[\mathbb{E}(Y_2)=\mu,\qquad
\operatorname{Var}(Y_2)=\sigma^2\mu^2.\]
Equivalently, the shape and scale parameters are \(1/\sigma^2\) and \(\sigma^2\mu\), respectively. The corresponding density and CDF are
\begin{equation}
\label{eq:gamma-margin}
f_2=f_2(y_2\mid\mu,\sigma)=
\frac{y_2^{1/\sigma^2-1}\exp\left(-\dfrac{y_2}{\sigma^2\mu}\right)}{\Gamma(1/\sigma^2)
(\sigma^2\mu)^{1/\sigma^2}}, \qquad
F_2 = F_2(y_2\mid\mu,\sigma), \qquad y_2>0.
\end{equation}

For the Gaussian copula, the copula dependence parameter is the latent correlation parameter \(\rho\). The copula is defined as
\begin{equation}
\label{eq:gaussian-copula}
C(u,v\mid\rho)
=\Phi_2\left(\Phi^{-1}(u),\Phi^{-1}(v)\mid\rho\right),\qquad
0\leq u,v\leq1,\quad -1<\rho<1,
\end{equation}
where \(\Phi(\cdot)\) is the standard univariate normal CDF and \(\Phi_2(\cdot,\cdot \mid\rho)\) is the standard bivariate normal CDF with correlation parameter \(\rho\). The partial derivative of the Gaussian copula with respect to its second argument is
\begin{equation}
\label{eq:gaussian-copula-derivative}
\frac{\partial C(u,v\mid\rho)}{\partial v}
=\Phi\left(\frac{\Phi^{-1}(u)-\rho\Phi^{-1}(v)}{
\sqrt{1-\rho^2}}\right).
\end{equation}

Combining the marginal quantities in Equations~\eqref{eq:poisson-margin}--\eqref{eq:gamma-margin}
with the Gaussian copula derivative in
Equation~\eqref{eq:gaussian-copula-derivative}, the mixed discrete--continuous log-likelihood contribution in Equation~\eqref{eq:log-likelihood} becomes
\begin{equation}
\label{eq:Poisson-Gamma-Gaussian-log}
\begin{split}
\ell(y_1,y_2\mid\lambda,\mu,\sigma,\rho)
={}&\log\Bigg[\Phi\left(\frac{\Phi^{-1}(F_1)-\rho\Phi^{-1}(F_2)
}{\sqrt{1-\rho^2}}\right)\\
&\hspace{2.2cm}-\Phi\left(
\frac{\Phi^{-1}(F_1^{-})-\rho\Phi^{-1}(F_2)}{
\sqrt{1-\rho^2}}\right)\Bigg] +\log f_2.
\end{split}
\end{equation}

When \(y_1=0\), \(F_1^{-}=0\). Because \(\Phi^{-1}(0)=-\infty\),
\[\Phi\left(\frac{\Phi^{-1}(F_1^{-})-\rho\Phi^{-1}(F_2)}{
\sqrt{1-\rho^2}}\right)=0.\]
Therefore, the second term inside the logarithm in
Equation~\eqref{eq:Poisson-Gamma-Gaussian-log} vanishes when \(y_1=0\).

For observation \(i\), the individual log-likelihood contribution is obtained by substituting
\[(y_1,y_2,\lambda,\mu,\sigma,\rho)=
(y_{1i},y_{2i},\lambda_i,\mu_i,\sigma_i,\rho_i)\]
into Equation~\eqref{eq:Poisson-Gamma-Gaussian-log}. Thus,
\[\ell_i=\ell(y_{1i},y_{2i}\mid\lambda_i,\mu_i,\sigma_i,\rho_i),\]
and the full log-likelihood is
\[\ell=\sum_{i=1}^{n}\ell_i.\]

\subsection*{Appendix B. Noncyclic boosting algorithm for distributional copula regression} \label{appb-algorithm}
\begin{algorithm}[H]
\label{alg:noncyclic-boosting}
\small
\DontPrintSemicolon
\caption{Noncyclic model-based boosting}

\SetKwInOut{Define}{Define}
\Define{Base learners $b_{k,j}^{(\bullet)}(x_j)$, where $k$ indexes the parameter, $j$ indexes the covariate, and $\bullet\in\{1,2,c\}$.\;}
\SetKwInOut{Set}{Set}
\Set{Step length, $s_{\mathrm{step}}\ll 1$. \; 
Maximum number of fitting iterations, $m_{\mathrm{stop}}$. \;
Split the observations into a training sample ($n_{\mathrm{train}}$) and an out-of-bag tuning sample ($n_{\mathrm{mstop}}$).\;}

\BlankLine
\textbf{(1)}
Initialize all additive predictors
$\widehat{\eta}_{k,[0]}^{(\bullet)}$ using suitable offset values\;
\BlankLine
\For{$m=1,\ldots,m_{\mathrm{stop}}$}{
    \ForEach{$\theta_k^{(\bullet)}\in\bm{\theta}$}{
        \textbf{(a)}
        Compute the parameter-specific negative gradient vector
        \[\begin{aligned}
        -\bm{g}_{k,[m]}^{(\bullet)}
        &= -\left(\left.\frac{\partial\omega
        \left(\bm{y}_i\mid\bm{\eta}(\bm{x}_i)\right)}{
        \partial\eta_k^{(\bullet)}}
        \right|_{\bm{\eta}=\widehat{\bm{\eta}}_{[m-1]}}
        \right)_{i=1}^{n_{\mathrm{train}}}.
        \end{aligned}\]
        
        \textbf{(b)}
        Fit each parameter-specific base learner
        $b_{k,j}^{(\bullet)}(x_j)$ to $-\bm{g}_{k,[m]}^{(\bullet)}$\;
        
        \textbf{(c)}
        Select the best-fitting base learner, $b_{k,j_k^\ast}^{(\bullet)}$, using the residual
        sum-of-squares criterion:
        \[j_k^\ast =
        \operatorname*{arg\,min}_{j\in\{1,\ldots,P_k^{(\bullet)}\}}
        \sum_{i=1}^{n_{\mathrm{train}}}
        \left[-g_{k,[m]}^{(\bullet)}(\bm{x}_i) -
        \widehat{b}_{k,j}^{(\bullet)}(x_{ij})\right]^2.\]
        \textbf{(d)}
        Compute the candidate empirical loss after the weak update:
        \[\begin{aligned}L_{k,[m]}
        =\sum_{i=1}^{n_{\mathrm{train}}}
        \omega\Big(\bm{y}_i\mid\,\widehat{\bm{\eta}}_{[m-1]}(\bm{x}_i)
        + s_{\mathrm{step}}\, \bm{e}_k
        \widehat{b}_{k,j_k^\ast}^{(\bullet)}(x_{ij_k^\ast})\Big),
        \end{aligned}\]
        where $\bm{e}_k$ is the unit vector corresponding to
        $\eta_k^{(\bullet)}$\;}
    \BlankLine
    \textbf{(2)}
    Select the parameter-specific predictor producing the smallest
    candidate loss:
    \[k^\ast = \operatorname*{arg\,min}_{k=1,\ldots,K} L_{k,[m]}.\]

    Update the selected additive predictor:
    \[\widehat{\eta}_{k^\ast,[m]}^{(\bullet)}(\bm{x}_i)
    = \widehat{\eta}_{k^\ast,[m-1]}^{(\bullet)}(\bm{x}_i) +
    s_{\mathrm{step}}\, \widehat{b}_{k^\ast,j_{k^\ast}^\ast}^{(\bullet)}
    (x_{ij_{k^\ast}^\ast}). \]

    \textbf{(3)}
    For all remaining predictors $k\neq k^\ast$, set
    \[\widehat{\eta}_{k,[m]}^{(\bullet)}(\bm{x}_i) =
    \widehat{\eta}_{k,[m-1]}^{(\bullet)}(\bm{x}_i).\]

    \textbf{(4)}
    Compute the out-of-bag risk:
    \[\operatorname{risk}_{\mathrm{oobag},[m]} =
    \sum_{i=1}^{n_{\mathrm{mstop}}}
    \omega\left(\bm{y}_i\mid\widehat{\bm{\eta}}_{[m]}(\bm{x}_i)\right).\]}
\BlankLine
\textbf{(5)}
Determine the optimal stopping iteration:
\[m_{\mathrm{stop}}^{\mathrm{opt}}=
\operatorname*{arg\,min}_{m\in\{1,\ldots,m_{\mathrm{stop}}\}}
\operatorname{risk}_{\mathrm{oobag},[m]}.\]
\end{algorithm}

\subsection*{Appendix C. Descriptive details of datasets} \label{appb-variables}

\begin{table}[!htbp]
\centering
\begin{threeparttable}
\caption{Variables in the synthetic telematics dataset}
\label{tab:VD2}
\small
\setlength{\tabcolsep}{6pt}
\renewcommand{\arraystretch}{1.08}
\begin{tabularx}{\linewidth}{@{}p{0.34\linewidth}>{\raggedright\arraybackslash}X@{}}
\toprule
\textbf{Variable} & \textbf{Description} \\\midrule

\rowcolor{black!6}[0pt][0pt]
\multicolumn{2}{@{}l@{}}{\textbf{Categorical covariates}} \\
\texttt{Insured.sex} & Sex of the insured driver: male or female \\
\texttt{Marital} & Marital status: single or married \\
\texttt{Car.use} & Vehicle use: private, commute, farmer, or commercial \\
\texttt{Region} & Region type: rural or urban \\
\texttt{Territory} & Territorial code: 55 distinct labels in $\{11,12,\ldots,91\}$ \\

\addlinespace[0.55em]
\rowcolor{black!6}[0pt][0pt]
\multicolumn{2}{@{}l@{}}{\textbf{Compositional covariates}\tnote{a}} \\
\texttt{Pct.drive.xxx} & Driving proportions by day of the week, where \texttt{xxx} is \texttt{mon}, \texttt{tue}, \texttt{wed}, \texttt{thr}, \texttt{fri}, \texttt{sat}, or \texttt{sun} \\
\texttt{Pct.drive.wkday},\newline\texttt{Pct.drive.wkend} & Driving proportions on weekdays and weekends \\

\addlinespace[0.55em]
\rowcolor{black!6}[0pt][0pt]
\multicolumn{2}{@{}l@{}}{\textbf{Percentage and ratio covariates}} \\
\texttt{Annual.pct.driven} & Annualized driving-day ratio: number of driving days divided by 365 \\
\texttt{Pct.drive.xhrs} & Driving proportion for durations exceeding $x$ hours, where $x\in\{2,3,4\}$ \\
\texttt{Pct.drive.rush.xx} & Driving proportion during morning or evening rush hours, where \texttt{xx} is \texttt{am} or \texttt{pm} \\

\addlinespace[0.55em]
\rowcolor{black!6}[0pt][0pt]
\multicolumn{2}{@{}l@{}}{\textbf{Continuous and integer covariates}} \\
\texttt{Duration} & Policy exposure, in years \\
\texttt{Insured.age} & Age of the insured driver, in years \\
\texttt{Car.age} & Age of the vehicle, in years \\
\texttt{Credit.score} & Credit score of the insured driver \\
\texttt{Annual.miles.drive} & Expected annual mileage declared by the insured driver \\
\texttt{Years.noclaims} & Number of years without claims \\
\texttt{Total.miles.driven} & Total distance driven, in miles \\
\texttt{Avgdays.week} & Average number of driving days per week \\
\texttt{Accel.xxmiles}\tnote{b} & Sudden acceleration events per 1,000 miles at threshold \texttt{xx}, in mph/s \\
\texttt{Brake.xxmiles}\tnote{b} & Sudden braking events per 1,000 miles at threshold \texttt{xx}, in mph/s \\
\texttt{Left.turn.intensityxx}\tnote{c} & Left-turn events per 1,000 miles at intensity level \texttt{xx} \\
\texttt{Right.turn.intensityxx}\tnote{c} & Right-turn events per 1,000 miles at intensity level \texttt{xx} \\

\addlinespace[0.55em]
\rowcolor{black!6}[0pt][0pt]
\multicolumn{2}{@{}l@{}}{\textbf{Response variables}} \\
\texttt{NB\_Claim} & Number of claims\\
\texttt{AMT\_Claim} & Aggregate claim amount \\
\bottomrule
\end{tabularx}

\begin{tablenotes}[flushleft]
\footnotesize
\item[a] Each compositional group sums to one (or 100\%): the seven day-of-week proportions form one group, and the weekday/weekend proportions form another.

\item[b] The acceleration and braking threshold labels are \texttt{06}, \texttt{08}, \texttt{09}, \texttt{11}, \texttt{12}, and \texttt{14}.
\item[c] The turning-intensity labels are \texttt{08}, \texttt{09}, \texttt{10}, \texttt{11}, and \texttt{12}.
\end{tablenotes}
\end{threeparttable}
\end{table}

\clearpage
\bibliographystyle{apalike}
\bibliography{JDCR}

\end{document}